\documentclass[10pt,twocolumn,letterpaper]{article}

\usepackage[pagenumbers]{cvpr} 

\usepackage{graphicx}
\usepackage{amsmath}
\usepackage{amssymb}
\usepackage{amsthm}
\usepackage{booktabs}
\usepackage{bm}
\usepackage{caption}
\newtheorem{theorem}{Theorem}

\newtheorem{lemma}[theorem]{Lemma}
\newtheorem{innercustomthm}{Theorem}
\newenvironment{customthm}[1]
{\renewcommand\theinnercustomthm{#1}\innercustomthm}
{\endinnercustomthm}
\usepackage[ruled,vlined]{algorithm2e}
\usepackage{caption}
\usepackage{subcaption}
\usepackage{multicol}
\usepackage{wrapfig, lipsum, booktabs}

\graphicspath{{figures/}}
\SetKwComment{Comment}{/* }{ */}

\usepackage[pagebackref,breaklinks,colorlinks]{hyperref}

\usepackage[capitalize]{cleveref}
\crefname{section}{Sec.}{Secs.}
\Crefname{section}{Section}{Sections}
\Crefname{table}{Table}{Tables}
\crefname{table}{Tab.}{Tabs.}

\def\cvprPaperID{5311} 
\def\confName{CVPR}
\def\confYear{2023}

\begin{document}

\title{Efficient Second-Order Plane Adjustment}

\author{Lipu ZHou\\
Meituan, Beijing, China\\
{\tt\small zhoulipu@meituan.com}
}
\maketitle

\begin{abstract}
   Planes are generally used in 3D reconstruction for depth sensors, such as RGB-D cameras and LiDARs. This paper focuses on the problem of estimating the optimal planes and sensor poses to minimize the point-to-plane distance. The resulting least-squares problem is referred to as plane adjustment (PA) in the literature, which is the counterpart of bundle adjustment (BA) in visual reconstruction. Iterative methods are adopted to solve these least-squares problems. Typically,  Newton’s method is rarely used for a large-scale least-squares problem, due to the high computational complexity of the Hessian matrix. Instead,  methods using an approximation of the Hessian matrix, such as the Levenberg-Marquardt (LM) method, are generally adopted. This paper challenges this ingrained idea. We adopt the Newton’s method to efficiently solve the PA problem. Specifically, given poses, the optimal planes have close-form solutions. Thus we can eliminate planes from the cost function, which significantly reduces the number of variables. Furthermore, as the optimal planes are functions of poses, this method actually ensures that the optimal planes for the current estimated poses can be obtained at each iteration, which benefits the convergence. The difficulty lies in how to efficiently compute the Hessian matrix and the gradient of the resulting cost. This paper provides an efficient solution.  Empirical evaluation shows that our algorithm converges significantly faster than the widely used LM algorithm. 
\end{abstract}

\section{Introduction}
\begin{figure}[t]
	\centering
	\includegraphics[width=0.9\linewidth]{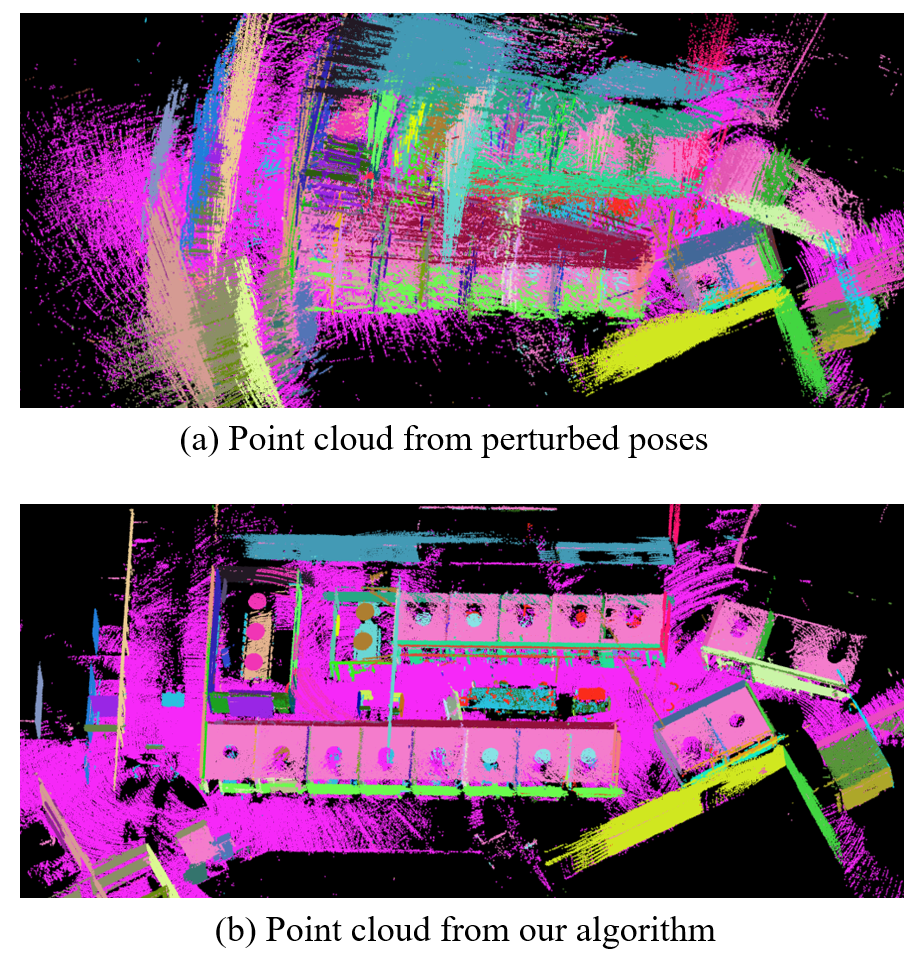}
	\caption{We use Gaussian noises to perturb the poses of dataset D in Fig.~\ref{fig:dataset}. The standard deviations for rotation and translation are $3^{\circ}$ and $0.3m$, respectively. The resulting point cloud (a) is in a mess. Fig. (b) shows the result from our algorithm. Our algorithm can quickly align the planes, as shown in Fig.~\ref{fig:iteration}.} \label{fig:demonstration} 
\end{figure} 
Planes ubiquitously exist in man-made environments, as demonstrated in Fig.~{\ref{fig:demonstration}}. Thus they are generally used in  simultaneous localization and mapping (SLAM) systems for depth sensors, such as RGB-D cameras \cite{kaess2015simultaneous,elghor2015planes,hsiao2017keyframe,kim2018linear,chen2022vip} and LiDARs \cite{zhang2014loam,liu2021balm,zhou2021lidar,zhou2021pi,zhou2022plc}. Just as bundle adjustment (BA) \cite{triggs1999bundle,agarwal2010bundle,zhou2020stochastic,demmel2021square} is important for visual  reconstruction \cite{agarwal2011building,mur2015orb,schonberger2016structure,campos2021orb}, jointly optimizing planes and depth sensor poses, which is called plane adjustment (PA) \cite{zhou2021lidar,zhou2021pi}, is critical for 3D reconstruction using depth sensors. This paper focuses on efficiently solving the large-scale PA problem.

The BA and PA problems both involve jointly optimizing  3D structures and sensor poses. As the two problems are similar, it is straightforward to apply the well-studied solutions for BA to PA, as done in \cite{zhou2020efficient,zhou2021pi}. However,  planes in PA can be eliminated, so that the cost function of the PA problem only depends on sensor poses, which significantly reduces the number of variables. This property provides a promising direction to efficiently solve the PA problem. However, it is difficult to compute the Hessian matrix and the gradient vector of the resulting cost. Although this problem was studied in several previous works \cite{ferrer2019eigen,liu2021balm}, no efficient solution has been proposed. This paper seeks to solve this problem. 

The main contribution of this paper is an efficient PA solution using Newton's method. We derive a closed-form solution for the Hessian matrix and the gradient vector for the PA problem whose computational complexity is  independent of the number of points on the planes. Our experimental results show that, in terms of the PA problem, Newton's method outperforms the widely-used Levenberg-Marquardt (LM) algorithm \cite{more1978levenberg} with Schur complement trick \cite{triggs1999bundle}.

\section{Related Work}

The PA problem is closely related to the BA problem.  In BA, points and camera poses are jointly optimized to minimize the reprojection error.  Schur complement \cite{triggs1999bundle,agarwal2010bundle,zhou2020stochastic} or the square root  method \cite{demmel2021squaresw,demmel2021square} is generally used to solve the  linear system of the iterative methods. The keypoint is to generate a reduced camera system (RCS) which only relates to camera poses. 

\begin{figure}[t]
	\centering
	\includegraphics[width=1\linewidth]{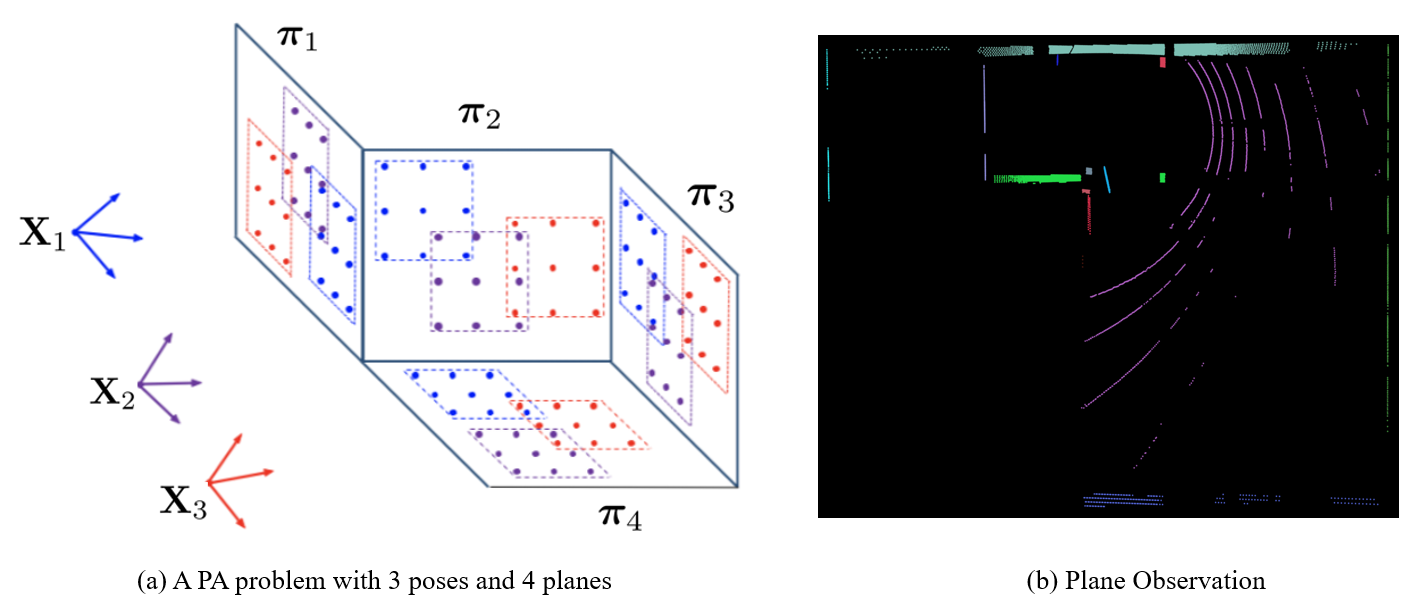}
	\caption{A schematic of the PA problem and the planes detected in a LiDAR scan. Unlike BA where each 3D point only has one observation, many points can be captured from a plane in PA. Assume that $N$ points are captured from a plane. The computational complexity of the Hessian matrix related to these points for BALM \cite{liu2021balm} is $O(N^{2})$. Thus, this method is infeasible for a large-scale problem. In contrast, the computational complexity of our algorithm is independent of $N$.} \label{fig:PA} 
\end{figure}

In PA, planes and poses are jointly optimized. Planes are the counterparts of points in BA. Thus, the well-known solutions for the BA problem can be applied to the PA problem \cite{zhou2020efficient,zhou2021lidar}.  In the literature, two cost functions are used to formulate the PA problem. The first one is the plane-to-plane distance which measures the difference between two plane parameters \cite{kaess2015simultaneous,hsiao2017keyframe}.  The value of the  plane-to-plane distance is related to the choice of the global coordinate system, which means  the selection of the global coordinate system will affect the accuracy of the results. The second one is the point-to-plane distance, whose value is invariant to choice of the global coordinate system. The solutions of different choices of the global coordinate system are equivalent up to a rigid transformation. Zhou \textit{et al.} \cite{zhou2020efficient} show that the point-to-plane distance can converge faster and lead to a more accurate result. But unlike  BA  where each 3D point  has only one 2D observation at a pose, a plane can generate many points at a pose as demonstrated in Fig.~\ref{fig:PA}. This means the point-to-plane distance probably leads to a very large-scale least-squares problem. Directly adopting the BA solutions is computationally infeasible for a large-scale PA problem. Zhou \textit{et al.} \cite{zhou2020efficient} propose to use the QR decomposition to accelerate the computation.

For a general least-squares problem with $M$ variables, the computational complexity of the Hessian matrix is $O(M^{2})$. Thus, in the computer vision community, it is ingrained that Newton's method is infeasible for a large-scale optimization problem, as calculating the Hessian matrix is  computationally demanding.  Instead, Gauss-Newton like iterative methods are generally adopted.  Suppose that $\mathbf{J}$ is the Jacobian matrix of the residuals. Gauss-Newton like methods actually approximates the Hessian matrix   by $\mathbf{H} \approx \mathbf{J}^{T}\mathbf{J}$. In theory, Newton's method can lead to a better quadratic approximation to the original cost function, which means the Newton's step probably yields a more accurate result. This in turn  may reduce the number of iterations for convergence.

The PA problem has a special property that the optimal plane parameters are determined by the poses. That is to say the point-to-plane cost actually only depends on the poses. This property is attractive, as it significantly reduces the number of variables, which makes using the Newton's method possible. Moreover, in the traditional framework, the correlation between the plane parameters and the poses are ignored.Thus, after one iteration, there is no guarantee that the new plane parameters are  optimal for the new poses. Using the property of the PA, it is possible to overcome this drawback, which may lead to faster convergence. Several previous works seek to exploit this property of PA.  Ferrer \cite{ferrer2019eigen} considered an algebraic point-to-plane distance and provided a closed-form gradient for the resulting cost. The algebraic cost may result in a suboptimal solution \cite{andrew2001multiple}, and the first-order optimization generally leads to slow convergence \cite{triggs1999bundle}.
Liu \textit{et al.} \cite{liu2021balm} provided analytic forms of the Hessian matrix and the gradient of the genuine point-to-plane cost. Assume that $N$ points  are captured from a plane at a pose.  The computational complexity of the Hessian matrix related to these points is $O(N^{2})$.  Since $N$ can be  large as shown in Fig. \ref{fig:PA}, this algorithm is  computationally demanding and infeasible for a large-scale problem. 

In summary, the potential benefits of the special property of the PA problem have not  been manifested in previous works. The bottleneck is how to efficiently compute the Hessian matrix and the gradient vector. This paper focuses on solving this problem.

\section{Problem Formulation}
In this paper we use italic, boldfaced lowercase and boldfaced uppercase letters to represent scalars, vectors and matrices, respectively. 

\subsection{Notations} \label{subsec:notations}

\textbf{Planes and Poses} \ A plane can be represented by a four-dimensional vector $\bm{\pi} = [\bm{n};d]$. We denote the rotational and translational components from a depth sensor coordinate system to the global coordinate system as $\mathbf{R} \in SO(3)$ and $\bm{t} \in \mathbb{R}^{3}$, respectively. To simplify the notation in the following description, we also use the following two matrices to represent a pose:
\begin{equation} \label{equ:pose}
	\mathbf{X} = \begin{bmatrix}
		\mathbf{R}, & \bm{t} \\ \mathbf{0}, &1
	\end{bmatrix} \in SE(3) \ \text{and} \ 
	\mathbf{T} = \begin{bmatrix}
		\mathbf{R}, & \bm{t} 
	\end{bmatrix}.
\end{equation}
As $\mathbf{R} \in SO(3)$, a certain parameterization is usually adopted in the optimization \cite{triggs1999bundle}. In this paper, we use the Cayley-Gibbs-Rodriguez (CGR) parameterization \cite{hesch2011direct}  to represent $\mathbf{R}$
\begin{equation} \label{equ:CGR}
	\mathbf{R}=\frac{{\mathbf{\bar{R}}}}{1+{{\bm{s}}^{T}}\bm{s}},\mathbf{\bar{R}}=\left( 1-{{\bm{s}}^{T}}\bm{s} \right){{\mathbf{I}}_{3}}+2{{\left[ \bm{s} \right]}_{\times }}+2\bm{s}{{\bm{s}}^{T}},
\end{equation}
where $\bm{s}=\left[s_{1};s_{2};s_{3}\right]$ is a three-dimensional vector. We adopt the CGR parameterization as it is a minimal representation for $\mathbf{R}$. Furthermore, unlike the angle-axis parameterization that is singular at $\mathbf{I}_{3}$, the CGR parameterization is well-defined at $\mathbf{I}_{3}$, and equals to $[0;0;0]$ which can accelerate the computation. We parameterize $\mathbf{T}$ as a six-dimensional vector $\bm{x} = [\bm{s}; \bm{t}]$.

\textbf{Newton's method} \ This paper adopts the damped Newton's method in the optimization. For a cost function $f(\bm{z})$, the damped Newton's method seeks to find its minimizer from an initial point. Assume that $\bm{z}_{n}$ is the solution at the $n$th iteration.  Given the Hessian matrix $\mathbf{H}_{f}(\bm{z}_{n})$ and the gradient $\bm{g}_{f}(\bm{z}_{n})$ at $\bm{z}_{n}$, $\bm{z}_{n}$ is updated by $\bm{z}_{n+1} = \bm{z}_{n} + \varDelta\bm{z}$. Here $\varDelta\bm{z}$ is from
\begin{equation} \label{equ:netwon}
	(\mathbf{H}_{f}(\bm{z}_{n})+\mu \mathbf{I}))\varDelta{z} = -\bm{g}_{f}(\bm{z}_{n}),
\end{equation}
where  $\mu$ is adjusted at each iteration  to make the value of $f(\bm{z})$ reduce, as done in the  LM algorithm \cite{more1978levenberg}. 

\textbf{Matrix Calculus} \ In the following derivation, we will use vector-by-vector,  vector-by-scalar, scalar-by-vector derivatives. Here we provide their definitions.

Assume $\bm{a} = [a_1; \cdots; a_N]\in \mathbb{R}^{N}$ is a vector function of $\bm{b} = [b_{1}, \cdots, b_{M}] \in \mathbb{R}^{M}$. The first-order partial derivatives of vector-by-vector  $\frac{\partial \bm{a}}{\partial \bm{b}}$, vector-by-scalar $\frac{\partial \bm{a}}{\partial b_{j}}$, and scalar-by-vector $\frac{\partial a_{i}}{\partial \bm{b}}$  are defined as
\begin{equation} \label{equ:vec_by_vec}
	\frac{\partial \bm{a}}{\partial \bm{b}} = \begin{bmatrix}
		\frac{\partial a_1}{\partial b_1} & \cdots & \frac{\partial a_N}{\partial b_1} \\
		\vdots & \ddots & \vdots \\
		\frac{\partial \bm{a}_{1}}{\partial b_M} & \cdots &\frac{\partial \bm{a}_{N}}{\partial \bm{b}_{M}}
	\end{bmatrix},
	\frac{\partial \bm{a}}{\partial b_{j}} = \begin{bmatrix}
		\frac{\partial a_{1}}{\partial b_{j}} \\
		\vdots \\
		\frac{\partial a_{N}}{\partial b_{j}}
	\end{bmatrix},
	\frac{\partial a_{i}}{\partial \bm{b}} = \begin{bmatrix}
		\frac{\partial a_{i}}{\partial b_{1}}\\
		\vdots \\
		\frac{\partial a_{i}}{\partial b_{M}}
	\end{bmatrix}
\end{equation}
where $\frac{\partial \bm{a}}{\partial \bm{b}}$ is an $M \times N$ matrix with $\frac{\partial a_{i}}{\partial b_{j}}$ as the $i$th row $j$th column element, $\frac{\partial \bm{a}}{\partial b_{i}}$ is an $N$-dimensional vector whose $i$th term is $\frac{\partial a_{i}}{\partial {b}_{j}}$, and $\frac{\partial a_{i}}{\partial \bm{b}}$ is an $M$-dimensional vector whose $i$th term is $\frac{\partial a_{i}}{\partial {b}_{i}}$.

\subsection{Optimal Plane Estimation} \label{subsec:Opt_plane} 
Given a set of $K$ points $\{\bm{p}_{i}\}$, the optimal plane $\hat{\bm{\pi}}$ can be estimated by minimizing the sum of squared point-to-plane distances
\begin{equation} \label{equ:plane_cost}
	\hat{\bm{\pi}} = \arg\min_{\bm{\pi}}\sum_{i}^{K}\left( \bm{n}^{T}\bm{p}_{i} + d\right) ^{2}, \ s.t. \ \left\| \bm{n} \right\| _2^{2} = 1.
\end{equation}
There is a closed-form solution for $\hat{\bm{\pi}}$. Let us define 
\begin{equation} \label{equ:Matrix_M}
	\small
	\mathbf{M} = \sum_{i=1}^{K}\left( \bm{p}_{i} - \bar{\bm{p}}\right)\left( \bm{p}_{i} - \bar{\bm{p}}\right)^{T} =  \mathbf{S} - K\bar{\bm{p}}\bar{\bm{p}}^{T}, 
\end{equation}
where $\mathbf{S} = \sum_{i=1}^{K} \bm{p}_{i}\bm{p}_{i}^{T}$ and $\bar{\bm{p}} = \frac{1}{K}\sum_{i}^{K}{\bm{p}}_{i}$. Assume that ${\lambda}_{3}(\mathbf{M})$ and $\bm{\xi}_{3}(\mathbf{M})$ are the smallest eigenvalue of $\mathbf{M}$ and the corresponding eigenvector, respectively. Using these notations, we can write the optimal plane $\hat{\bm{\pi}} = [\hat{\bm{n}}; \hat{d}]$ as
\begin{equation} \label{equ:opt_pi}
	\hat{\bm{n}} = \bm{\xi}_{3}(\mathbf{M}), \ \hat{d} = -\hat{\bm{n}}^{T} \bar{\bm{p}}.
\end{equation}
Furthermore, the cost of (\ref{equ:plane_cost}) at $\hat{\bm{\pi}}$ equals to ${\lambda}_{3}(\mathbf{M})$, \textit{i.e.,}
\begin{equation} \label{equ:opt_pi_cost}
	\begin{split}
		{\lambda}_{3}(\mathbf{M}) = \sum_{i=1}^{K}\left( \hat{\bm{n}}\bm{p}_{i} + \hat{d}\right) ^{2} = \min_{\bm{\pi}}\sum_{i=1}^{K}\left( {\bm{n}}\bm{p}_{i} + {d}\right) ^{2}.
	\end{split}
\end{equation} 
The above property will be used to eliminate planes in PA.

\subsection{Plane Adjustment}
Assume that there are $M$ planes and $N$ poses. According to section \ref{subsec:notations}, the $i$th  plane can be represented by a four-dimensional vector $\bm{\pi}_{i} = [\bm{n}_{i};d_{i}]$. The $j$th pose is denoted as $\bm{x}_{j}$. The observation of $\bm{\pi}_{i}$ at $\bm{x}_{j}$ is a set of $N_{ij}$ points $\mathbb{P}_{ij} = \{\bm{p}_{ijk} \in \mathbb{R}^{3}\}_{k=1}^{N_{ij}}$.  For a 3D point $\bm{p}_{ijk}$, we use $\tilde{\bm{p}}_{ijk}= [\bm{p}_{ijk};1]$ to represent the homogeneous coordinates of $\bm{p}_{ijk}$.  Then, the transformation  from the local coordinate system at $\bm{x}_{j}$ to the global coordinate system can be represented as 
\begin{equation} \label{equ:g_coor}
	\bm{p}^{g}_{ijk} = \mathbf{R}_{j}\bm{p}_{ijk} + \bm{t}_{j} = \mathbf{T}_{j}\tilde{\bm{p}}_{ijk},
\end{equation} 
where $\mathbf{T}_{j}$ is defined in (\ref{equ:pose}).
Then the distance $d_{ijk}$ from $\bm{p}_{ijk}$ to $\bm{\pi}_{i}$ has the form
\begin{equation} \label{equ:pt_2_pi_dis}
	\begin{split}
		d_{ijk}(\bm{\pi}_{i},\mathbf{x}_{j}) & = \bm{n}_{i}^{T}\left( \mathbf{R}_{j}\bm{p}_{ijk} + \bm{t}_{j} \right) + d_{i}\\
		& = \bm{n}_{i}^{T}\mathbf{T}_{j}\tilde{\bm{p}}_{ijk} + d_{j} = \bm{\pi}^{T}_{i}\tilde{\bm{p}}_{ijk}^{g}.
	\end{split}
\end{equation}
The PA problem is to jointly adjust the $M$ planes $\left\lbrace \bm{\pi}_{i} \right\rbrace $ and the $N$ sensor poses $\left\lbrace \bm{x}_{j} \right\rbrace $  to minimize the sum of squared point-to-plane distances. Specifically, using  (\ref{equ:pt_2_pi_dis}), we can formulate the cost function of the PA problem as


\begin{equation}\label{equ:PA_cost}
	\small
	\min_{\left\lbrace \bm{\pi}_{i}\right\rbrace, \left\lbrace \mathbf{x}_{j}\right\rbrace} \sum_{i=1}^{N}\underbrace{\sum_{ j \in obs(\bm{\pi}_i)}\sum_{k=1}^{N_{ij}} d_{ijk}^{2}(\bm{\pi}_{i},\mathbf{x}_{j})}_{C_{i}\left(\bm{\pi}_{i}, \mathbb{X}_{i}\right), \ \mathbb{X}_{i} = \{\bm{x}_{j}| j\in obs(\bm{\pi}_i)\}} = \min_{\{\bm{\pi}_{i}\}, \{\bm{x}_{j}\}}\sum_{i=1}^{N_{i}}C_{i}\left(\bm{\pi}_{i}, \mathbb{X}_{i}\right).
\end{equation}
where $obs(\bm{\pi}_i)$ represents the indexes of poses where $\bm{\pi}_{i}$ can be observed, and $C_{i}(\bm{\pi}_{i},\mathbb{X}_{i})$ accumulates the errors from  $N_{i} = \sum_{j \in obs(\bm{\pi}_{i}) }N_{ij}$ points captured at the set of poses $\mathbb{X}_{i}$.  According to (\ref{equ:Matrix_M}) and (\ref{equ:g_coor}), we get
\begin{equation} \label{equ:M_i}
	\small
	\begin{split}
		\mathbf{M}_{i}(\mathbb{X}_{i}) &= \sum_{j \in obs(\bm{\pi}_i)} \mathbf{S}_{ij} - N_{i}\bar{\bm{p}}_{i}\bar{\bm{p}}_{i}^{T}, 
	\end{split}
\end{equation}
where  $\bar{\bm{p}}_{i}  = \frac{1}{N_i}\sum_{j\in obs(\bm{\pi}_i)}\sum_{k=1}^{N_{ij}} \bm{p}_{ijk}^{g}$ and $\mathbf{S}_{ij} = {\sum_{k=1}^{N_{ij}} \bm{p}_{ijk}^{g}\left( {\bm{p}_{ijk}^{g}}\right) ^{T}}$.	Here the elements in $\mathbf{M}_{i}$, $\mathbf{S}_{ij}$ and $\bar{\bm{p}}_{i}$ in (\ref{equ:M_i}) are all functions of the poses in  $\mathbb{X}_{i}$. Substituting $\bm{p}_{ijk}^{g}$ in (\ref{equ:g_coor}) into $\mathbf{S}_{ij}$ and $\bar{\bm{p}}_{i}$ in (\ref{equ:M_i}), we have
\begin{equation} \label{equ:S_p}
	\small
	\begin{split}
	\mathbf{S}_{ij} &= \mathbf{T}_{j}\underbrace{\sum_{k=1}^{N_{ij}}\tilde{\bm{p}}_{ijk}\tilde{\bm{p}}_{ijk}^{T}}_{\mathbf{U}_{ij}}\mathbf{T}_{j}^{T} = \mathbf{T}_{j}\mathbf{U}_{ij}\mathbf{T}_{j}^{T}, \\
	\bar{\bm{p}}_{i} & = \frac{1}{N_i}\sum_{j \in obs(\bm{\pi}_{i})}\mathbf{T}_{j}\underbrace{\sum_{k=1}^{N_{ij}} \tilde{\bm{p}}_{ijk}}_{\tilde{\bm{p}}_{ij}} = \frac{1}{N_i}\sum_{j\in obs(\bm{\pi}_{i})}\mathbf{T}_{j}\tilde{\bm{p}}_{ij}.
	\end{split}
\end{equation}
Here $\mathbf{U}_{ij}$ and $\tilde{\bm{p}}_{ij}$ in (\ref{equ:S_p}) are constants. We only need to compute them once, and reuse them in the iteration. 

According to (\ref{equ:opt_pi}), given poses in $\mathbb{X}_{i}$, the optimal  solution for $\bm{\pi}_{i}$ has a closed-form expression  $\hat{\bm{\pi}}_{i} =[ \hat{\bm{n}}_{i};\hat{d}_{i}]$, where $ \hat{\bm{n}}_{i} =\bm{\xi}_{3}(\mathbf{M}_{i}(\mathbb{X}_{i}))$ and $\hat{d}_{i} = -\hat{\bm{n}}\bar{\bm{p}}_{i}$.  As $\mathbf{M}_{i}$ and $\bar{\bm{p}}_{i}$ are functions of the poses in $\mathbb{X}_{i}$, $\hat{\bm{\pi}}_{i}$ is also a function of the poses in $\mathbb{X}_{i}$. That is to say $\hat{\bm{\pi}}_{i}$ is completely determined by the  poses in $\mathbb{X}_{i}$. To simplify the notation, let us define
\begin{equation}
	\lambda_{i,3}(\mathbb{X}_{i}) = \lambda_{3}(\mathbf{M}_{i}(\mathbb{X}_{i})),
\end{equation}
which represents the smallest eigenvalue of $\mathbf{M}_{i}(\mathbb{X}_{i})$.

Substituting the optimal plane estimation $\hat{\bm{\pi}}_{i}$ into $C_{i}(\bm{\pi}_{i}, \mathbb{X}_{i})$ in (\ref{equ:PA_cost}) and using (\ref{equ:opt_pi_cost}), we have
\begin{equation} \label{equ:min_pi}
	\lambda_{i,3}(\mathbb{X}_{i})  =  C_{i}\left( \hat{\bm{\pi}}_{i}, \mathbb{X}_{i}\right).
\end{equation}
Using (\ref{equ:min_pi}), we can formulate the PA problem in (\ref{equ:PA_cost}) as
\begin{equation} \label{equ:PA_cost_pose}
	\{\hat{\mathbf{x}}_{j}\} = \arg\min_{\{\mathbf{X}_{j}\}}\bm{\tau}, \ \bm{\tau} = \sum_{i=1}^{M} \lambda_{i,3}(\mathbb{X}_{i}).
\end{equation}
The cost function (\ref{equ:PA_cost_pose}) only depends on the sensor poses, which significantly reduces the number of variables. However, as it is the sum of squared point-to-plane distances, we cannot apply the widely used Gauss-Newton-like methods to minimize it, where the Jacobian matrix of  residuals are required. Here we adopt the Newton's method to solve it. The crux for applying the Newton's method to minimize (\ref{equ:PA_cost_pose}) is how to compute the gradient and the Hessian matrix  of (\ref{equ:PA_cost_pose}) efficiently.
In the following sections, we provide a closed-form solution for them. To simplify the notation, we omit the variables of functions in the following description (\textit{e.g.}, $\lambda_{i,3}(\mathbb{X}_{i}) \rightarrow \lambda_{i,3}$).

\section{Newton's Iteration for Plane Adjustment}
Let us denote the gradient and the Hessian matrix of $\tau$ in (\ref{equ:PA_cost_pose}) as $\bm{g}$ and $\mathbf{H}$, and denote the 6-dimensional gradient vector for $\bm{x}_{j}$ as $\bm{g}_{j}$ and the $6 \times 6$ Hessian matrix block for $\bm{x}_{j}$ and $\bm{x}_{k}$ as $\mathbf{H}_{jk}$ (note that here $i$ can equal to $j$). Then $\bm{g}$ and $\mathbf{H}$ can be written in the block form as $\bm{g}=(\bm{g}_{j}) \in \mathbb{R}^{6N}$ and $\mathbf{H} = (\mathbf{H}_{jk}) \in \mathbb{R}^{6N \times 6N}$.
%
%

The $i$th plane $\bm{\pi}_{i}$ is observed by the poses in $\mathbb{X}_{i}$.  Assume $j$th pose  $\bm{x}_{j} \in \mathbb{X}_{i}$ and  the $k$th pose $\bm{x}_{k} \in \mathbb{X}_{i}$.  Let us define
\begin{equation} \label{equ:g_ij and H_ijk}
	\bm{g}_{j}^{i} = \frac{\partial \lambda_{i,3}}{\partial \bm{x}_{j}}, \
	\mathbf{H}_{jk}^{i} = \frac{\partial^{2} \lambda_{i,3}}{\partial \bm{x}_{j} \partial \bm{x}_{k}}.
\end{equation}
According to (\ref{equ:PA_cost_pose}), we have
\begin{equation} \label{equ:g_j and H_jk}
	\bm{g}_{j} = \sum_{i \in \mathbb{P}_{j}} \bm{g}^i_{j}, \ \mathbf{H}_{jk} = \sum_{i \in \mathbb{P}_{jk}} \mathbf{H}_{jk}^{i},
\end{equation}
where $\mathbb{P}_{j}$ is the set of  planes observed by $\bm{x}_{j}$, and  $\mathbb{P}_{jk}$ is 
the set of planes observed by $\bm{x}_{j}$ and $\bm{x}_{k}$ simultaneously. If $j = k$, here $\mathbb{P}_{jk}$ equals to $\mathbb{P}_{j}$.  From (\ref{equ:g_j and H_jk}), we know that the key point to get $\bm{g}$ and $\mathbf{H}$ is to compute $\bm{g}_{j}^{i}$ and $\mathbf{H}_{jk}^{i}$ in (\ref{equ:g_ij and H_ijk}).

\subsection{Partial Derivatives of Eigenvalue}

According to (\ref{equ:g_ij and H_ijk}), $\lambda_{i,3}$ is a function of poses in $\mathbb{X}_{i}$, and $\bm{x}_{j} \in \mathbb{X}_{i}$ and $\bm{x}_{k} \in \mathbb{X}_{i}$.  Assume that $x_{jm}$ and $x_{kn}$ are the $m$th and $n$th elements of $\bm{x}_{j}$ and $\bm{x}_{k}$, respectively.  
In this section, we first consider the first-order partial derivation $\frac{\partial\lambda_{i,3}}{\partial x_{jm}}$ and the second-order partial derivation $\frac{\partial^{2}\lambda_{i,3}}{\partial x_{jm} \partial x_{kn}}$. 

$\lambda^{3}_{i,3}$ is a root of the equation $|\mathbf{M}_{i}(\mathbb{X}_{i}) - \lambda_{i}\mathbf{I}_{3}| = 0$, where $|\cdot|$ denotes the determinant of a matrix. Assume  $m_{ef}$ is the $e$th row $f$th column term of $\mathbf{M}_{i}(\mathbb{X}_{i})$.  $|\mathbf{M}_{i}(\mathbb{X}_{i}) - \lambda_{i}\mathbf{I}_{3}| = 0$ is a  cubic equation with the following form   
\begin{equation} \label{equ:cub}
	-\lambda^{3}_{i,3} + a_{i}\lambda^{2}_{i,3} + b_{i}\lambda_{i,3} + c_{i} = 0,
\end{equation} 
where $a_{i} = m_{11} + m_{22} + m_{33}$, $b_{i} = m_{12}^2 + m_{13}^2 + m_{23}^2 - m_{11}m_{22}- m_{11}m_{33} - m_{22}m_{33}$, and $c_{i} = - m_{33}m_{12}^2 + 2m_{12}m_{13}m_{23} - m_{22}m_{13}^2 - m_{11}m_{23}^2 + m_{11}m_{22}m_{33}$. Here $a_{i}$, $b_{i}$ and $c_{i}$ are all functions of the poses in $\mathbb{X}_{i}$. It is known that the root of a cubic equation has a closed form. One solution to compute $\frac{\partial\lambda_{i,3}}{\partial x_{jm}}$ and $\frac{\partial^{2}\lambda_{i,3}}{\partial x_{jm} \partial x_{kn}}$ is to directly differentiate the root. However, the formula of the root is too complicated. Here we introduce a simple way to compute them. Briefly, we employ the implicit function theorem \cite{krantz2002implicit} to compute them. Let us define
\begin{equation}
	\bm{\chi}_{i} = \begin{bmatrix}
		\lambda_{i,3}^{2} \\  \lambda_{i,3} \\ 1 
	\end{bmatrix}, \
	\bm{\eta}_{i} =\begin{bmatrix}
			a_{i} \\
			b_{i} \\
			c_{i}
	\end{bmatrix}
	\bm{\kappa}_{i} = \begin{bmatrix}
		-3 \\ 2a_{i} \\ b_{i}
	\end{bmatrix}.
\end{equation}
$\frac{\partial\lambda_{i,3}}{\partial x_{jm}}$ and $\frac{\partial^{2}\lambda_{i,3}}{\partial x_{jm} \partial x_{kn}}$ are presented in Lemma \ref{lemma:first_order} and \ref{lemma:second_order}. The proofs of the following lemmas and theorems are in the\textbf{ supplementary material}.

\begin{lemma} \label{lemma:first_order}
	$\frac{\partial\lambda_{i,3}}{\partial x_{jm}}$ has a closed-form expression as
	\begin{equation}  \label{equ:first_order}
		\frac{\partial\lambda_{i,3}}{\partial x_{jm}} = - \varphi_{i}  \bm{\delta}_{jm}^{i} \cdot \bm{\chi}_{i},
	\end{equation}
	where $\cdot$ represents the dot product and $\varphi_{i} = \left( \bm{\kappa}_{i} \cdot \bm{\chi}_{i}\right)^{-1}$ and $\bm{\delta}_{jm}^{i} = \frac{\partial \bm{\eta}_{i}}{\partial x_{jm}}$.
\end{lemma}

\begin{lemma} \label{lemma:second_order}
	$\frac{\partial^2{}\lambda_{i,3}}{\partial x_{jm}\partial x_{kn}}$ has a closed-form expression as
	\begin{equation} \label{equ:second_order}
		\small
		\frac{\partial^{2}\lambda_{i,3}}{\partial x_{jm}\partial x_{kn}} = -\varphi_{i}\left(\bm{\delta}_{jm}^{i} \cdot \frac{\partial \bm{\chi}_{i}}{\partial x_{kn}} + \bm{\chi}_{i} \cdot \frac{\partial \bm{\delta}_{jm}^{i}}{\partial x_{kn}} - \frac{\partial \lambda_{i,3}}{\partial x_{jm}} \frac{\partial \varphi_{i}^{-1}}{\partial x_{kn}}\right).
	\end{equation}
\end{lemma}
Let us define
\begin{equation} \label{equ:d_abc}
	\small
	\begin{split}
		&\bm{\alpha}_{j}^{i} = \frac{\partial a_{i}}{\partial \bm{x}_{j}},  \bm{\beta}_{j}^{i} = \frac{\partial b_{i}}{\partial \bm{x}_{j}}, \bm{\gamma}_{j}^{i} = \frac{\partial c_{i}}{\partial \bm{x}_{j}}, \mathbf{\Delta}_{j}^{i} = [\bm{\alpha}_{j}^{i},\bm{\beta}_{j}^{i},\bm{\gamma}_{j}^{i}],\\ 
		&\bm{\alpha}_{k}^{i} = \frac{\partial a_{i}}{\partial \bm{x}_{k}}, \bm{\beta}_{k}^{i} = \frac{\partial b_{i}}{\partial \bm{x}_{k}}, \bm{\gamma}_{k}^{i} = \frac{\partial c_{i}}{\partial \bm{x}_{k}},  \mathbf{\Delta}_{k}^{i} = [\bm{\alpha}_{k}^{i},\bm{\beta}_{k}^{i},\bm{\gamma}_{k}^{i}],\\
		&\mathbf{H}^{a_{i}}_{jk} = \frac{\partial^{2} a_{i}}{\partial \bm{x}_{j} \partial \bm{x}_{k}},
		\mathbf{H}^{b_{i}}_{jk} = \frac{\partial^{2} b_{i}}{\partial \bm{x}_{j} \partial \bm{x}_{k}},
		\mathbf{H}^{c_{i}}_{jk} = \frac{\partial^{2} c_{i}}{\partial \bm{x}_{j} \partial \bm{x}_{k}}.
	\end{split}
\end{equation}
Using the above lemmas and notations, we can derive $\bm{g}_{j}^{i}$ and $\mathbf{H}_{jk}^{i}$.
\begin{customthm}{1} \label{theorem:g_H}
	$\bm{g}_{j}^{i}$ and $\mathbf{H}_{jk}^{i}$  have the forms
	\begin{equation} \label{equ:g_H}
		\begin{split}
			\bm{g}_{j}^{i} &= -\varphi_{i}\bm{\Delta}_{j}^{i}\bm{\chi}_{i},  \\
			\mathbf{H}_{jk}^{i} &= \varphi_{i}\left(\mathbf{K}_{jk}^{i} - {\lambda_{3,i}^{2}}\mathbf{H}^{a_{i}}_{jk} - \lambda_{3,i}\mathbf{H}^{b_{i}}_{jk} - \mathbf{H}^{c_{i}}_{jk}\right),
		\end{split}
	\end{equation}
	where $\mathbf{K}_{jk}^{i} = \bm{g}_{j}^{i}{(\bm{u}_{k}^{i})}^{T} -  \bm{v}_{j}^{i}({\bm{g}_{k}^{i}})^{T}$, $\bm{u}_{k}^{i} = 2\lambda_{i,3}\bm{\alpha}_{j}^{i}+\bm{\beta}_{j}^{i} + (2a-6\lambda_{i,3})\bm{g}_{k}^{i}$, $\bm{v}_{j}^{i} = 2\lambda_{i,3}\bm{\alpha}_{k}^{i}+\bm{\beta}_{k}^{i}$, and similar to $\bm{g}_{j}^{i}$, $\bm{g}_{k}^{i}=-\varphi_{i}\bm{\Delta}_{k}^{i}\bm{\chi}_{i}$ is the gradient block for $\bm{x}_{k}$.
\end{customthm}

The formula of $\mathbf{H}_{jk}^{i}$ in (\ref{equ:g_H}) is applicable to the case that $j = k$. From Theorem \ref{theorem:g_H}, we know that the key point to get $\bm{g}_{j}^{i}$ and $\mathbf{H}_{jk}^{i}$ is to get the  derivatives of $a_{i}$, $b_{i}$ and $c_{i}$ in (\ref{equ:d_abc}).

\subsection{Partial Derivatives of $a_{i}$, $b_{i}$ and $c_{i}$}

As shown in (\ref{equ:cub}), $a_{i}$, $b_{i}$ and $c_{i}$ are functions of the elements in $\mathbf{M}_{i}$. Using this relationship, we can easily derive the partial derivatives in (\ref{equ:d_abc}). For instance, as $a_{i} = m_{11} + m_{22} + m_{33}$, we have
\begin{equation}
	\frac{\partial a_{i}}{\partial \bm{x}_{j}} = \frac{\partial m_{11}}{\partial \bm{x}_{j}} + \frac{\partial m_{22}}{\partial \bm{x}_{j}} + \frac{\partial m_{33}}{\partial \bm{x}_{j}}.
\end{equation}
Thus, to get the first- and second-order partial derivatives of $a_{i}$, $b_{i}$ and $c_{i}$ with respect to $\bm{x}_{j}$ and $\bm{x}_{k}$ in (\ref{equ:d_abc}), we need to derive the form of $\mathbf{M}_{i}$ in terms of  $\bm{x}_{j}$ and $\bm{x}_{k}$.

\begin{lemma} \label{lemma:mean_pt_decomp}
	In terms of $\bm{x}_{j}$ and $\bm{x}_{k}$, $\bar{\bm{p}}_{i}$ in (\ref{equ:S_p}) has the form
	\begin{equation}
		\bar{\bm{p}}_{i}(\bm{x}_{j}, \bm{x}_{k}) =  \mathbf{T}_{j}\bm{q}_{ij} + \mathbf{T}_{k}\bm{q}_{ik} + \bm{c}_{ijk},
	\end{equation}
	where $\bm{q}_{ij} = \frac{1}{N_{i}}\bm{\tilde{p}}_{ij}$, $\bm{q}_{ik} = \frac{1}{N_{i}}\bm{\tilde{p}}_{ik}$, and $\bm{c}_{ijk} =  \frac{1}{N_i}\sum_{n\in \mathbb{O}_{jk}}\mathbf{T}_{n}\tilde{\bm{p}}_{in}$. Here  $\mathbb{O}_{jk}$ represents the set of poses where $\bm{\pi}_{i}$ can be observed, excluding the $\bm{j}$th and $\bm{k}$th poses (\textit{i.e.}, $\mathbb{O}_{jk} = obs(\bm{\pi}_{i}) - \{j,k\}$).
	
	In terms of $\bm{x}_{j}$, $\bar{\bm{p}}_{i}$  has the form
	\begin{equation}
		\bar{\bm{p}}_{i}(\bm{x}_{j}) = \mathbf{T}_{j}\mathbf{q}_{ij} + \bm{c}_{ij},
	\end{equation}
	where $\bm{c}_{ij} = \mathbf{T}_{k}\mathbf{q}_{ik}+\bm{c}_{ijk}$.
\end{lemma}

Using Lemma \ref{lemma:mean_pt_decomp}, we can have the following theorem for $\mathbf{M}_{i}$ in (\ref{equ:M_i}).
\begin{customthm}{2} \label{theorem:M_i}
	In terms of $\bm{x}_{j}$, $\mathbf{M}_{i}$ in (\ref{equ:M_i}) can be written as
	\begin{equation} \label{equ:Mij}
		\mathbf{M}_{i}(\bm{x}_{j}) = \mathbf{T}_{j}\mathbf{Q}_{j}^{i}\mathbf{T}_{j}^{T} + \mathbf{T}_{j}\mathbf{K}_j^{i} + (\mathbf{K}_j^{i})^{T}\mathbf{T}_{j}^{T} + \mathbf{C}_{j}^{i},
	\end{equation}
	where $\mathbf{Q}_{j}^{i} = \mathbf{S}_{ij} - N_{j}\bm{q}_{ij}\bm{q}_{ij}^{T}$ and $\mathbf{K}_{j}^{i} = -N_{i}\bm{q}_{ij}\bm{c}_{ij}^T$.
	
	In terms of $\bm{x}_{j}$ and $\bm{x}_{k}$, $\mathbf{M}_{i}$ can be written as
	\begin{equation}\label{equ:M_ijk}
		\mathbf{M}_{i}(\bm{x}_{j},\bm{x}_{k}) = \mathbf{T}_{j}\mathbf{O}_{jk}^{i}\mathbf{T}_{k}^{T}  +  \mathbf{T}_{k}{(\mathbf{O}_{jk}^{i})}^{T}\mathbf{T}_{j}^{T} + \mathbf{C}_{jk}^{i}.
	\end{equation}
	where $\mathbf{O}_{jk}^{i} = -N_{i}\bm{q}_{ij}\bm{q}_{ik}^{T}$.
\end{customthm}

Here we do not provide the detailed formulas for $\mathbf{C}_{j}^{i}$ and $\mathbf{C}_{jk}^{i}$, as they will be eliminated in the partial derivative. Actually, only $\mathbf{Q}_{j}^{i}$, $
\mathbf{K}_{j}^{i}$, and $\mathbf{O}_{jk}^{i}$ are required to compute the partial derivatives in (\ref{equ:d_abc}).  Equation (\ref{equ:Mij}) is used to compute the first- and second-order partial derivatives of $a_{i}$, $b_{i}$ and $c_{i}$ with respect to $\bm{x}_{j}$. Equation (\ref{equ:M_ijk}) is used to compute the second-order partial derivatives of $a_{i}$, $b_{i}$ and $c_{i}$ with respect to $\bm{x}_{j}$ and $\bm{x}_{k}$.

\subsection{Efficient Iteration}
From Theorem \ref{theorem:M_i}, we can easily derive the elements of $\mathbf{M}_{i}(\bm{x}_{j})$ and $\mathbf{M}_{i}(\bm{x}_{j}, \bm{x}_{i})$.  Specifically, each element of them is a second-order polynomial in terms of the elements in $\mathbf{T}_{j}$ and  $\mathbf{T}_{k}$. Assume 
$m_{ef}(\bm{x}_{j})$ and $m_{ef}(\bm{x}_{j}, \bm{x}_{k})$ are the $e$th row $f$th column element of $\mathbf{M}_{i}(\bm{x}_{j})$ and $\mathbf{M}_{i}(\bm{x}_{j},\bm{x}_{k})$, respectively. $m_{ef}(\bm{x}_{j})$ and $m_{ef}(\bm{x}_{j}, \bm{x}_{k})$ are linear combinations of monomials with respect to the elements in $\mathbf{T}_{j}$ and $\mathbf{T}_{k}$.  Substituting (\ref{equ:CGR}) into $m_{ef}(\bm{x}_{j})$ and $m_{ef}(\bm{x}_{j}, \bm{x}_{k})$, we have
\begin{equation}
	\begin{split}
		m_{ef}(\bm{x}_{j}) & = \bm{c}_{ef} \cdot \bm{h}_{ef}(\bm{x}_{j}), \\
		m_{ef}(\bm{x}_{j}, \bm{x}_{k}) & = \bm{d}_{ef} \cdot \bm{g}_{ef}(\bm{x}_{j}, \bm{x}_{k}),
	\end{split}
\end{equation}
where $\bm{c}_{ef}$  is  determined by $\mathbf{Q}_{j}^{i}$ and $\mathbf{K}^{i}_{j}$ in (\ref{equ:Mij}), $\bm{d}_{ef}$  is  determined by $\mathbf{O}_{jk}^{i}$ in (\ref{equ:M_ijk}), and $\bm{h}_{ef}$ and $\bm{g}_{ef}$ are two vector functions.
Let us first consider the first-order partial derivative of $m_{ef}(\bm{x}_{j})$ with respect to $\bm{x}_{j}$. It has the form
\begin{equation} \label{equ:d_h}
	\frac{\partial m_{ef}(\bm{x}_{j})}{\partial \bm{x}_{j}} = \frac{\partial \bm{h}_{ef}(\bm{x}_{j})}{\partial \bm{x}_{j}}  \bm{c}_{ef},
\end{equation}
where the vector-by-vector derivative $\frac{\partial \bm{h}_{ef}(\bm{x}_{j})}{\partial \bm{x}_{j}}$ is defined in (\ref{equ:vec_by_vec}).
To efficiently compute (\ref{equ:d_h}), we consider a special pose $\mathbf{T}_{0} = [\mathbf{R}_{0}, \bm{t}_{0}]$ where $\mathbf{R}_{0} = \mathbf{I}_{3}$ and $\bm{t}_{0} = [0;0;0]$. Let us denote the parameterization of $\mathbf{T}_{0}$   as $\bm{x}_{0}$. As the CGR parameterization defined in (\ref{equ:CGR})  for $\mathbf{I}_{3}$ is $[0;0;0]$, we have $\bm{x}_{0} = [0;0;0;0;0;0]$. For $\bm{x}_{j} = \bm{x}_{0}$, the matrix $\left. \frac{\partial \bm{h}_{ef}(\mathbf{x}_{j})}{\partial \bm{x}_{j}}\right| _{{\mathbf{x}_{j}} = \mathbf{x}_{0}}$ has many zero terms. Similarly, the second-order partial derivatives of $\bm{h}_{ef}(\bm{x}_{j})$ and $\bm{g}_{ef}(\bm{x}_{j})$ at $\bm{x}_{j} = \bm{x}_{0}$ and $\bm{x}_{k} = \bm{x}_{0}$ are simple. As $\bm{h}_{ef}(\bm{x}_{j})$ and $\bm{g}_{ef}(\bm{x}_{j},\bm{x}_{k})$ only depends on $\bm{x}_{j}$ and $\bm{x}_{k}$, we can compute their partial derivatives at $\bm{x}_{0}$ once, and then reuse them during the iteration.  Here we introduce a method to make the iteration stay at $\mathbf{x}_{0}$ for each pose.

Assume that $\{\mathbf{X}_{j}^{n}\}$ are the poses after the $n$th iteration. Then we can update $\mathbf{U}_{ij}$ and $\tilde{\bm{p}}_{ij}$ in (\ref{equ:S_p}) by
\begin{equation} \label{equ:update_U_p}
	\mathbf{U}_{ij}^{n+1} = \mathbf{X}_{j}^{n}\mathbf{U}_{ij}( \mathbf{X}_{j}^{n})^{T} \ \text{and} \ \tilde{\bm{p}}_{ij}^{n+1} = \mathbf{X}_{j}^{n}\tilde{\bm{p}}_{ij}.
\end{equation}
Substituting $\mathbf{U}_{ij}^{n+1}$ and $\tilde{\bm{p}}_{ij}^{n+1}$ into (\ref{equ:M_i}), we  get a new matrix $\mathbf{M}_{i}(\mathbb{X}_{i})^{n+1}$, which can finally lead to a new cost $\tau^{n+1}$ in (\ref{equ:PA_cost_pose}). As  the points have been transformed by $\{\mathbf{X}_{j}^{n}\}$,  each pose should start with $\mathbf{X}_0$ for $\tau^{n+1}$. Assume that $\Delta \bm{x}^{n+1}_{j}$ is the result  from the $(n+1)$th iteration for the $j$th pose. We can obtain the corresponding transformation matrix $\Delta\mathbf{X}_{j}^{n+1}$ using (\ref{equ:CGR}).  Then we can update $\mathbf{X}_{j}^{n}$ by
\begin{equation}
	\mathbf{X}_{j}^{n+1} = \Delta\mathbf{X}_{j}^{n+1}\mathbf{X}_{j}^{n} 
\end{equation}
Furthermore, the update steps in  (\ref{equ:update_U_p}) will not introduce additional computation. This is because the damped Newton's method requires to compute the cost $\tau$ in (\ref{equ:PA_cost_pose}) to adjust $\mu$ in (\ref{equ:netwon})  after each iteration, which  will perform the computation in  (\ref{equ:update_U_p}).

\begin{figure*} 
	\centering
	\includegraphics[width=0.8\linewidth]{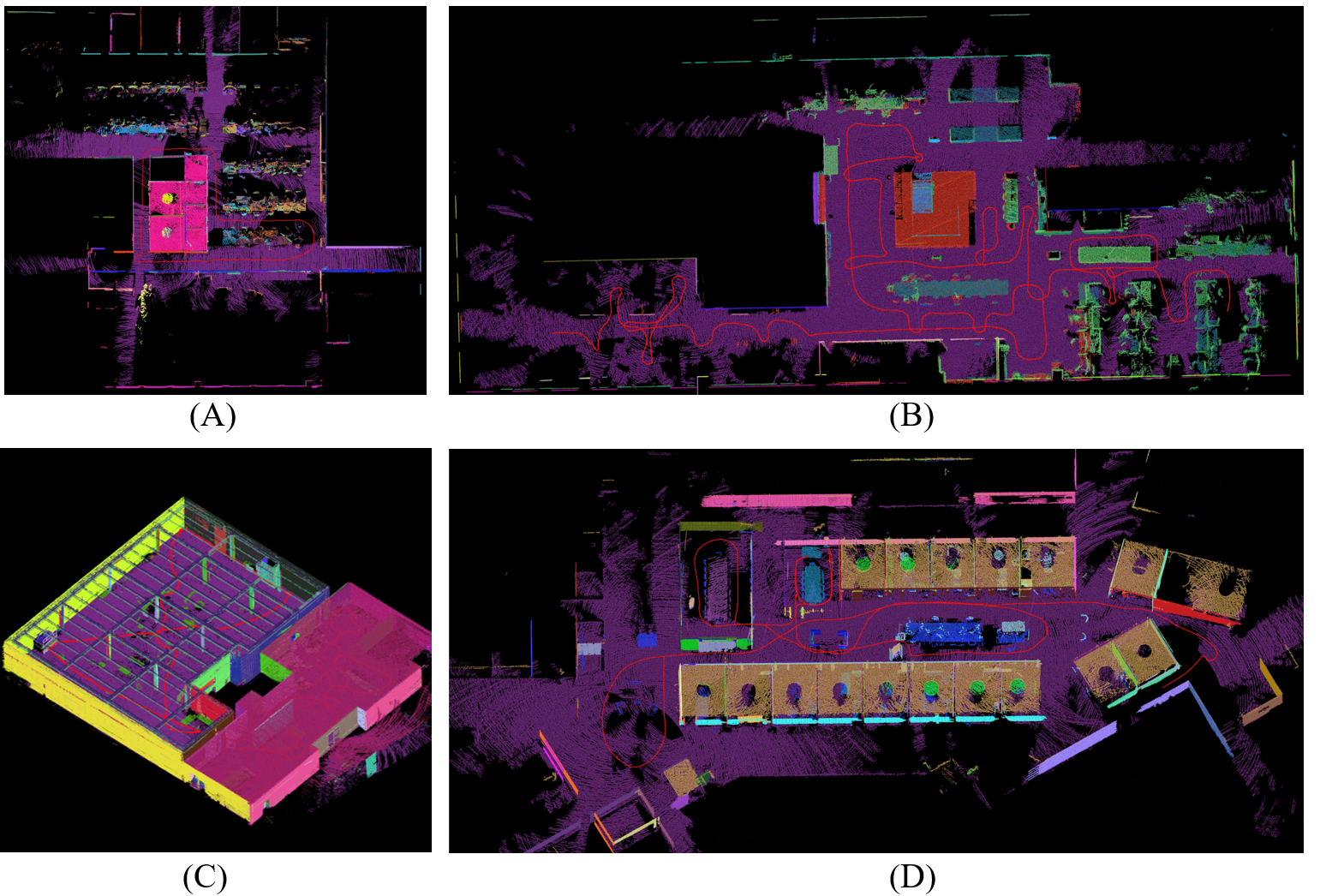}
	\caption{The four datasets used in this paper. The four datasets have 472, 1355, 1606, 1184 poses, respectively. Roofs are removed to show the trajectories.} \label{fig:dataset} 
\end{figure*}

\begin{figure*} 
	\centering
	\includegraphics[width=0.8\linewidth]{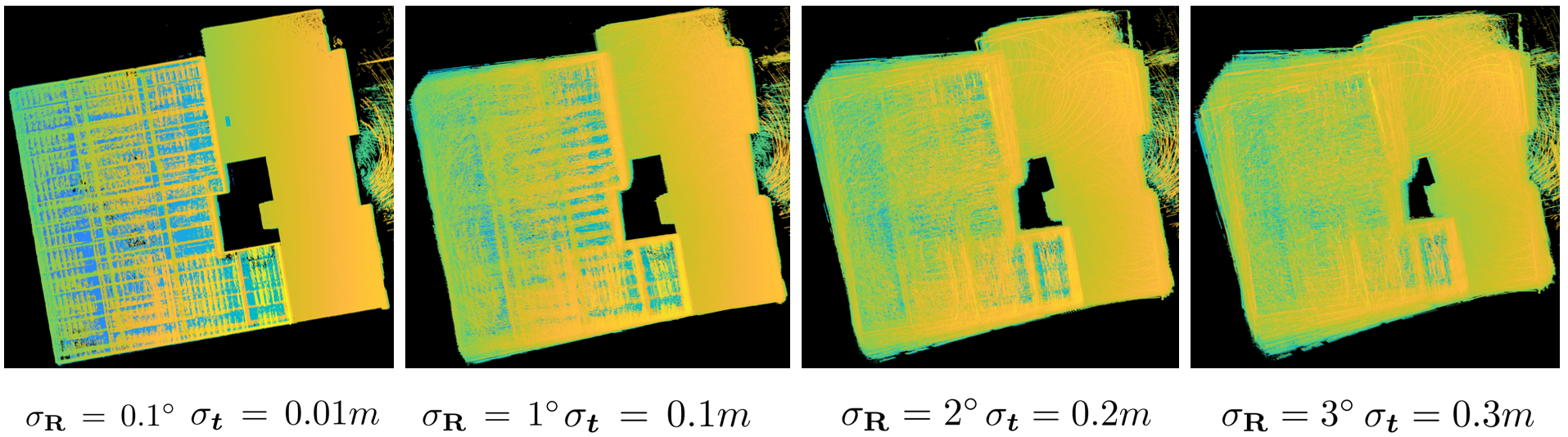}
	\caption{The point clouds of dataset C after the poses were perturbs by the four noise levels. } \label{fig:noisy_pc} 
\end{figure*}

\begin{figure*}
	\centering
	\includegraphics[width=0.93\linewidth]{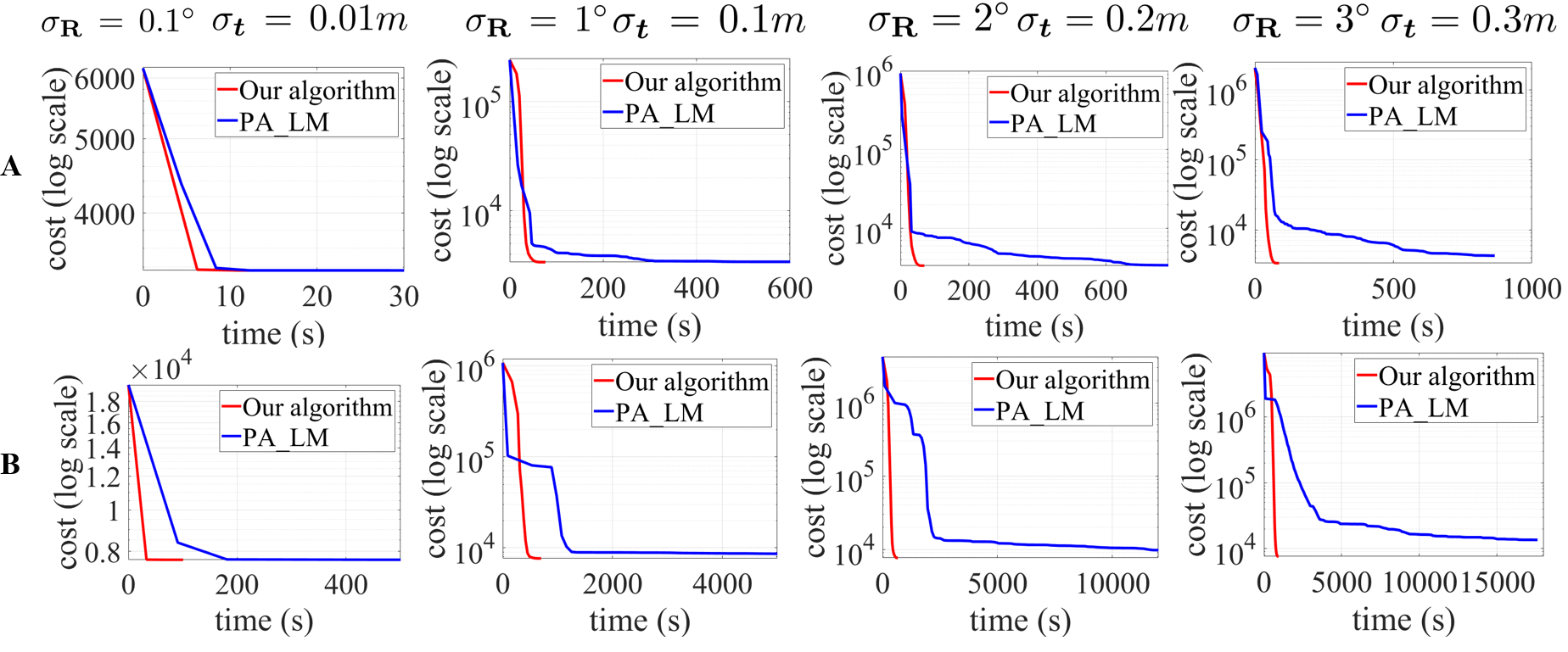}
	\includegraphics[width=0.93\linewidth]{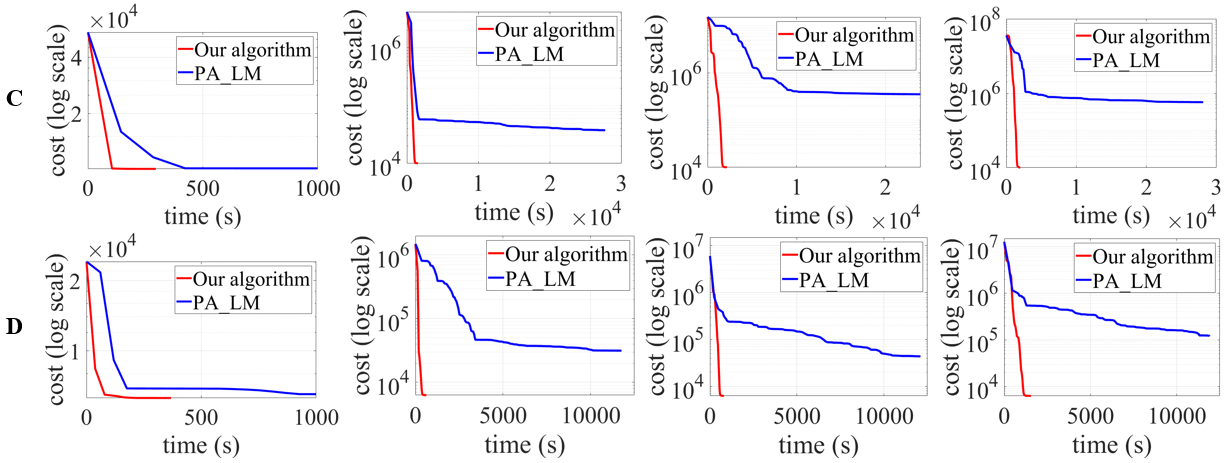}
	\caption{The results of our algorithm and PA\_LM \cite{zhou2020efficient} under different noise levels. It is clear that our algorithm converges significantly faster than PA\_LM.} \label{fig:resultts} 
\end{figure*}

\begin{figure}
	\centering
	\includegraphics[width=0.9\linewidth]{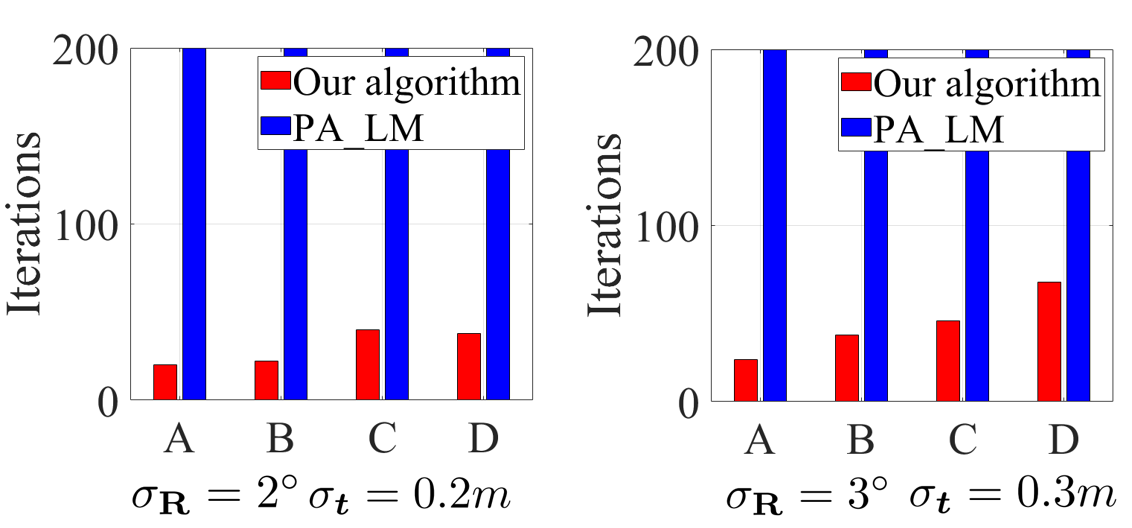}
	\caption{The number of iterations for our algorithm and PA\_LM \cite{zhou2020efficient}. Our algorithm requires much fewer number of iterations than PA\_LM. } \label{fig:iteration}  
\end{figure}

\subsection{Algorithm Summary}

We first construct $\mathbf{H}$ and $\bm{g}$. For each plane $\bm{\pi}_{i}$, we solve the cubic equation (\ref{equ:cub}), and select  the smallest root $\lambda_{i,3}$. For $\bm{x}_{j}$, we construct $\mathbf{M}(\bm{x}_{j})$ in (\ref{equ:Mij}), and calculate the partial derivatives of $a_{i}$, $b_{i}$ and $c_{i}$ with respect to $\bm{x}_{j}$ in (\ref{equ:d_abc}). Then, we use (\ref{equ:g_H}) to compute $\bm{g}_{j}^{i}$ and $\mathbf{H}_{jj}^{i}$ and use (\ref{equ:g_j and H_jk})  to update $\bm{g}_{j}$ and $\mathbf{H}_{jj}$. For $\bm{x}_{j}$ and $\bm{x}_{k}$, $\mathbf{M}_{i}(\bm{x}_{j}, \bm{x}_{k})$ is generated, and then the partial derivatives of $a_{i}$, $b_{i}$ and $c_{i}$ with respect to $\bm{x}_{j}$ and $\bm{x}_{j}$ in (\ref{equ:d_abc}) are computed. Then, $\mathbf{H}_{jk}^{i}$ can be easily obtained from (\ref{equ:g_H}), and $\mathbf{H}_{jk}$ in (\ref{equ:g_j and H_jk}) is updated accordingly. Using $\mathbf{H}$ and $\bm{g}$, we conduct the damped Newton's step in (\ref{equ:netwon}). After each iteration, $\mathbf{U}_{ij}$ and $\tilde{\bm{p}}_{ij}$ are updated by (\ref{equ:update_U_p}). The proposed algorithm is summarized in Algorithm \ref{alg:2nd_pa}. Let us denote the mean and variance of the number of observations per plane as $\bar{K}$ and $\sigma^{2}$, respectively. According to \cite{demmel2021square}, the computational complexity of the Hessian matrix is $O(M(\bar{K}^{2} + \sigma^{2} ))$, which is of the same order as the Schur complement trick.

\begin{algorithm} 
	\caption{Second-order plane adjustment for $N$ poses and $M$ planes}\label{alg:2nd_pa}
	\While{not converge}{
		$\mathbf{H} = zeros(6N,6N)$, $\bm{g} = zeros(6N,1)$; \\
		\For{$i \in [1, M]$}{
			{ \Comment{\small Compute $\bm{g}$ and the diagonal terms of $\mathbf{H}.$}}
			\For{$j \in obs(\bm{\pi}_{i})$}{ 
				Compute $\mathbf{M}_{i}(\bm{x}_{j})$ using (\ref{equ:Mij}); \\
				Compute $\bm{\alpha}_{j}^{i}$, $\bm{\beta}_{j}^{i}$, $\bm{\gamma}_{j}^{i}$, $\mathbf{H}_{jj}^{a_{i}}$, $\mathbf{H}_{jj}^{b_{i}}$, $\mathbf{H}_{jj}^{c_{i}}$ using (\ref{equ:d_abc}); \\
				Compute  $\mathbf{H}_{jj}^{i}$ and $\bm{g}_{j}^{i}$ using (\ref{equ:g_H});\\
				$\mathbf{H}_{jj} = \mathbf{H}_{jj} + \mathbf{H}_{jj}^{i}$, $\bm{g}_{j} = \bm{g}_{j} + \bm{g}_{j}^{i} $;\\
			}	
			\Comment{\small Compute other terms of $\mathbf{H}.$}
			\For{$j \in obs(\bm{\pi}_{i})$}{
				\For{$k \in obs(\bm{\pi}_{i})$ and $k > j$}{
					Compute $\mathbf{H}_{jk}^{a_{i}}$, $\mathbf{H}_{jk}^{b_{i}}$, $\mathbf{H}_{jk}^{c_{i}}$ using (\ref{equ:d_abc});\\
					Compute  $\mathbf{H}_{jk}^{i}$ using (\ref{equ:g_H});\\
					$\mathbf{H}_{jk} = \mathbf{H}_{jk} + \mathbf{H}_{jk}^{i}$; \\
					$\mathbf{H}_{kj} = \mathbf{H}_{kj} + (\mathbf{H}_{jk}^{i})^{T}$; \\
				}
			}
		}
		Conduct the damped Newton's step in (\ref{equ:netwon}) ; \\
		Update $\mathbf{U}_{ij}$ and $\tilde{\bm{p}}_{ij}$ using (\ref{equ:update_U_p});
	}
\end{algorithm}

\section{Experiments}

\subsection{Setup}
In this section, we evaluate the performance of our algorithm and the traditional LM algorithm \cite{zhou2020efficient} (PA\_LM). We obtain the c++ code of PA\_LM  from the author of \cite{zhou2020efficient}, which was implemented by the Ceres library \cite{Agarwal_Ceres_Solver_2022}. Our damped Newton's method is implemented according to the implementation of the LM algorithm in Ceres. The damped Newton's method and the LM algorithm are with the same parameters. Specifically, the  initial value of the damping factor $\mu$ in (\ref{equ:netwon}) is set to $10^{-4}$. The maximum number of iterations is set to 200, and the early stopping tolerances (such as the cost value change  and the norm of gradient) are set to $10^{-7}$.
All the experiments were conducted on a desktop with an Intel i7 cpu and 64G memory.

\subsection{Datasets}
We collected four datasets using a VLP-16 LiDAR. We used the LiDAR SLAM algorithm \cite{zhou2021lidar} to detect the planes and establish the plane association. Fig.~\ref{fig:dataset} shows the four datasets. Similar to the evaluation of  BA algorithms \cite{agarwal2010bundle,zhou2020stochastic,demmel2021square}, we perturb the pose, and compare the convergence speed of different PA algorithms. Specifically, we directly add Gaussian noises to the translation, and randomly generate an angle-axis vector from a Gaussian distribution to perturb the rotation. After the poses are perturb, we use (\ref{equ:opt_pi}) to get the initial plane parameters for PA\_LM \cite{zhou2020efficient}.

We evaluate the performance of different algorithms under different noise levels. Let us denote the standard deviation (std) of the Gaussian noises for rotation and translation as $\sigma_{\mathbf{R}}$ and $\sigma_{\bm{t}}$, respectively. We consider four noise levels: $\sigma_{\mathbf{R}} = 0.1^{\circ}$ and $\sigma_{\bm{t}} = 0.01 m$, $\sigma_{\mathbf{R}} = 1^{\circ}$ and $\sigma_{\bm{t}} = 0.1 m$, $\sigma_{\mathbf{R}} = 2^{\circ}$ and $\sigma_{\bm{t}} = 0.2 m$, and $\sigma_{\mathbf{R}} = 3^{\circ}$ and $\sigma_{\bm{t}} = 0.3 m$. Fig.~\ref{fig:noisy_pc} demonstrates the point clouds of dataset C after the poses are perturbed by the four noise levels.

\subsection{Results}
Fig.~\ref{fig:resultts} and Fig.~\ref{fig:iteration} illustrates the results. It is clear that our algorithm converges faster than PA\_LM.  PA\_LM works well at small noises (such as $\sigma_{\mathbf{R}} = 0.1^{\circ}$ and $\sigma_{\bm{t}} = 0.01 m$). As the noise level increases, PA\_LM tends to converge slower. For dataset A and B, PA\_LM does not converge before the maximum number of iterations reaches when  $\sigma_{\mathbf{R}} = 3^{\circ}$ and $\sigma_{\bm{t}} = 3 m$. For dataset C and D, the performance of PA\_LM gets worse. It converges very slowly after the noise level reaches $\sigma_{\mathbf{R}} = 1^{\circ}$ and $\sigma_{\bm{t}} = 0.1 m$. In contrast, the impact of the noise level on our algorithm is small. Our algorithm is more robust to the noise. 

\subsection{Discussion}
 Our algorithm converges faster than  PA\_LM for all the noise levels. This is because our algorithm takes advantage of the special relationship between planes and poses in (\ref{equ:opt_pi}). This does not only significantly reduce the number of variables, but only  ensures planes obtain the optimal estimation with respect to the current pose estimation after each iteration. Although PA\_LM jointly optimizes planes and poses, it cannot guarantee planes achieve the optimal value after each iteration.  That is to say even if our algorithm and PA\_LM get the same poses, our algorithm can reach a smaller cost. Thus our algorithm converges faster.

\section{Conclusion}
In the computer vision community, Newton's method is generally considered too expensive for a large-scale least-squares problem. This paper adopts the Newton's method to efficiently solve the PA problem. Our algorithm takes advantage of the fact that the optimal planes are determined by the poses, so that the number of unknowns can be significantly reduced. Furthermore, this property can ensure to obtain the optimal planes when we update the poses. The difficulty lies in how to efficiently compute the Hessian matrix and the gradient vector. The key contribution of this paper is to provide a closed-form solution for them. The experimental results show that our algorithm can converge faster than the LM algorithm. 

\section{Supplementary Material}

\subsection{Implicit Function Theorem}

Here we introduce the implicit function theorem \cite{ift}. We used it to derive Lemma \ref{lemma:first_order}.

\noindent
\textbf{Implicit Function Theorem} \ \textit{Let $\bm{f}:\mathbb{R}^{n+m} \rightarrow \mathbb{R}^{m}$ be a continuously differentiable function, and let $ \mathbb {R}^{n+m}$ have coordinates $\left[ {\bm{x}},{\bm{y}}\right]$. Fix a point $\left[\bm{a},\bm{b} \right] = \left[a_{1}, \cdots a_{n}, b_{1} \cdots b_{m} \right]$ with $f(\bm{a},\bm{b}) = \mathbf{0}$, where $\mathbf{0} \in \mathbb{R}^{m}$ is the zero vector. If the Jacobian matrix of $f$ with respect to $\bm{y}$ is invertible at $\left[ \bm{a},\bm{b}\right] $, then there exists an open set $\mathbb{U} \subset \mathbb{R}^{n}$ containing $\bm{a}$ such that there exists a unique continuously differentiable function $\bm{g}:\mathbb{U} \rightarrow \mathbf{R}^{m}$ such that $\bm{g}(\bm{a}) = \bm{b}$, and $\bm{f}(\bm{x},\bm{g}(\bm{x}))=0$ for all $\bm{x} \in \mathbb{U}$. Moreover, the Jacobian matrix of $\bm{g}$ in $\mathbb{U}$ with respect to $\bm{x}$ is given by the matrix product:
	\begin{equation}
		\frac{\partial \bm{g}}{\partial \bm{x}}(\bm{x}) = -{\mathbf{J}_{\bm{f},\bm{y}}(\bm{x},\bm{g}(\bm{x}))}^{-1}\mathbf{J}_{\bm{f},\bm{x}}(\bm{x},\bm{g}(\bm{x}))
	\end{equation}
	where ${\mathbf{J}_{\bm{f},\bm{y}}(\bm{x},\bm{g}(\bm{x}))}$ is the Jacobian matrix of $\bm{f}$ with respect to $\bm{y}$ at $\left[ \bm{x}, \bm{g}(\bm{x})\right] $, and ${\mathbf{J}_{\bm{f},\bm{x}}(\bm{x},\bm{g}(\bm{x}))}$ is the Jacobian matrix of $\bm{f}$ with respect to $\bm{x}$ at $\left[ \bm{x}, \bm{g}(\bm{x})\right] $.}

\textbf{From the implicit function theorem, we know that we can get $\frac{\partial \bm{g}}{\partial \bm{x}}\left( \bm{x}\right) $ without knowing the exact form of $\bm{g}(\bm{x})$}.

\subsection{Proof of Lemma \ref{lemma:first_order}}
\begin{proof}
	$a_{i}$, $b_{i}$, $c_{i}$ are functions of poses in $\mathbb{X}_{i}$.  Here we only consider one variable $x_{jm}$ of $\bm{x}_{j}$ (\textit{i.e.}, the $m$th entry of $\bm{x}_{j} \in \mathbb{X}_{i}$). To compute $\frac{\partial \lambda_{i,3}}{\partial x_{jm}}$, we treat $x_{jm}$ as the only unknown and other variables in $\mathbb{X}_{i}$  as constants. Thus, $a_{i}$, $b_{i}$, $c_{i}$ are functions of $x_{jm}$. Then, we define
	\begin{equation}
		f(x_{jm},\lambda_{i,3}) = -\lambda^{3}_{i,3} + a_{i}\lambda^{2}_{i,3} + b_{i}\lambda_{i,3} + c_{i}.
	\end{equation}
	Then we have
	\begin{equation} \label{equ:ift_part}
		\begin{split}
			\frac{\partial f}{\partial \lambda_{i,3}} &= -3\lambda_{i,3}^{2} + 2a_{i}\lambda_{i,3} + b_{i}, \\
			\frac{\partial f}{\partial x_{jm}} &= \lambda_{i,3}^{2}\frac{\partial a_{i}}{\partial x_{jm}} + \lambda_{i,3}\frac{\partial b_{i}}{\partial x_{jm}} + \frac{\partial c_{i}}{\partial x_{jm}}.
		\end{split}
	\end{equation}
	According to the definition of $\bm{\delta}_{jm}^{i}$ in (\ref{equ:first_order}), it has the form
	\begin{equation} \label{equ:delta_jm}
		\bm{\delta}_{jm}^{i} = \frac{\partial \bm{\eta}_{j}}{\partial x_{jm}} =  \begin{bmatrix}
			\frac{\partial a_{i}}{{x}_{jm}} \\
			\frac{\partial b_{i}}{x_{jm}} \\
			\frac{\partial c_{i}}{x_{jm}}
		\end{bmatrix}
	\end{equation}
	Substituting the definitions of $\varphi_{i}$, $\bm{\chi}_{i}$, $\bm{\kappa}_{i}$, and $\bm{\delta}_{jm}$ into (\ref{equ:ift_part}), we have
	\begin{equation} \label{equ:ift_replace}
		\begin{split}
			\frac{\partial f}{\partial \lambda_{i,3}} &= \bm{\kappa}_{i} \cdot \bm{\chi}_{i} = \varphi_{i}^{-1}, \\
			\frac{\partial f}{\partial x_{jm}} &= \bm{\delta}_{jm} \cdot \bm{\chi}_{i}.
		\end{split}
	\end{equation}
	Using the implicit function theorem, for $f(x_{jm},\lambda_{i,3}) = 0$, we have
	\begin{equation} \label{equ:ift_lambda_x}
		\frac{\partial \lambda_{i,3}}{\partial x_{jm}} = -\frac{\frac{\partial f}{\partial x_{jm}}}{\frac{\partial f}{\partial \lambda_{i,3}}}
	\end{equation}
	Substituting (\ref{equ:ift_replace}) into (\ref{equ:ift_lambda_x}), we finally get
	\begin{equation}
		\frac{\partial \lambda_{i,3}}{\partial x_{jm}}  = -\varphi_{i}\bm{\delta}_{jm}^{i}\cdot\bm{\chi}_{i}.
	\end{equation}
\end{proof}

\subsection{Proof of Lemma \ref{lemma:second_order}}

\begin{proof}
	
	We first compute the partial derivative of $\frac{\partial \lambda_{i,3}}{\partial x_{jm}}$ in (\ref{equ:first_order}) with respect to $x_{kn}$. According to the production rule of calculus, we have
	\begin{equation} \label{equ:2nd_order_derivation}
		\frac{\partial^2 \lambda_{i,3}}{\partial x_{jm} \partial x_{km}} = -\varphi_{i}\bm{\delta}_{jm}^{i} \cdot \frac{\partial \bm{\chi}_{i}}{\partial x_{kn}} -\varphi_{i}\bm{\chi}_{i} \cdot \frac{\partial \bm{\delta}_{jm}^{i}}{\partial x_{kn}} - \bm{\delta}_{jm}^{i} \cdot \bm{\chi}_{i}\frac{\partial \varphi_{i}}{\partial x_{kn}}.
	\end{equation}
	
	Let us first focus on the term $\frac{\partial \varphi_{i}^{-1}}{\partial x_{kn}}$ in (\ref{equ:2nd_order_derivation}). As  $\varphi_{i} = \left(\bm{\kappa}_{i} \cdot \bm{\chi}_{i} \right)$, we have
	\begin{equation} \label{equ:d_phi}
		\frac{\partial \varphi_{i}}{\partial x_{kn}} = -\left(\bm{\kappa}_{i} \cdot \bm{\chi}_{i} \right)^{-2}\frac{\partial \left(\bm{\kappa}_{i} \cdot \bm{\chi}_{i} \right)}{\partial x_{kn}} = -\varphi_{i}^{2}\frac{\partial \varphi_{i}^{-1} }{\partial x_{kn}}.
	\end{equation} 
	Now let us consider $\bm{\delta}_{jm}^{i} \cdot \bm{\chi}_{i}\frac{\partial \varphi_{i}}{\partial x_{kn}}$ that is the third term in (\ref{equ:2nd_order_derivation}). Using (\ref{equ:d_phi}), we have
	\begin{equation} \label{equ:3rd_term}
		\begin{split}
			\bm{\delta}_{jm}^{i} \cdot \bm{\chi}_{i}\frac{\partial \varphi_{i}}{\partial x_{kn}} & = -\bm{\delta}_{jm}^{i} \cdot \bm{\chi}_{i}\varphi_{i}^{2}\frac{\partial \varphi_{i}^{-1} }{\partial x_{kn}} \\
			& = -\varphi_{i}\underbrace{\left(\varphi_{i}\bm{\delta_{jm}^{i}} \cdot\bm{\chi}_{i} \right)}_{\frac{\partial\lambda_{i,3}}{\partial x_{jm}}} \frac{\partial \varphi_{i}^{-1} }{\partial x_{kn}} \\
			& = -\varphi_{i}\frac{\partial\lambda_{i,3}}{\partial x_{jm}}\frac{\partial \varphi_{i}^{-1} }{\partial x_{kn}}
		\end{split}
	\end{equation}
	Substituting (\ref{equ:3rd_term}) into (\ref{equ:2nd_order_derivation}), we have
	\begin{equation}
		\small
		\begin{split}
			\frac{\partial^2 \lambda_{i,3}}{\partial x_{jm} \partial x_{km}} &= -\varphi_{i}\bm{\delta}_{jm}^{i} \cdot \frac{\partial \bm{\chi}_{i}}{\partial x_{kn}} -\varphi_{i}\bm{\chi}_{i} \cdot \frac{\partial \bm{\delta}_{jm}^{i}}{\partial x_{kn}} + \varphi_{i}\frac{\partial\lambda_{i,3}}{\partial x_{jm}}\frac{\partial \varphi_{i}^{-1} }{\partial x_{kn}} \\
			&= -\varphi_{i}\left(\bm{\delta}_{jm}^{i} \cdot \frac{\partial \bm{\chi}_{i}}{\partial x_{kn}} + \bm{\chi}_{i} \cdot \frac{\partial \bm{\delta}_{jm}^{i}}{\partial x_{kn}} - \frac{\partial \lambda_{i,3}}{\partial x_{jm}} \frac{\partial \varphi_{i}^{-1}}{\partial x_{kn}}\right)
		\end{split}
	\end{equation}
\end{proof}

\subsection{Proof of Theorem \ref{theorem:g_H}}

\begin{figure} 
	\centering
	\includegraphics[width=1\linewidth]{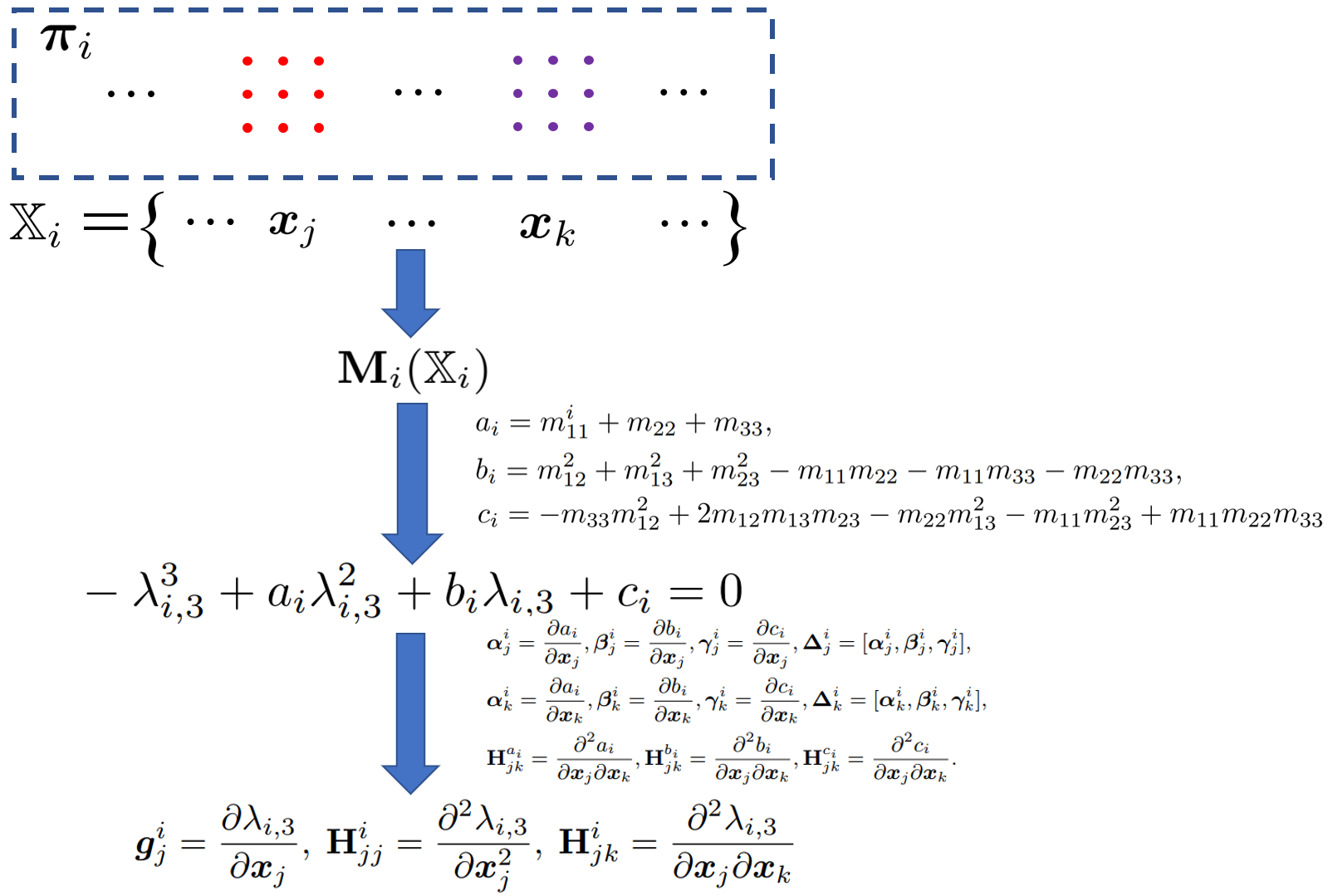}
	\caption{Summary of our algorithm. $\mathbb{X}_{i}$ is the set of poses which can see $\bm{\pi}_{i}$. Here $\bm{x}_{j} \in \mathbb{X}_{i}$ and $\bm{x}_{k} \in \mathbb{X}_{i}$. The key point of our algorithm is to get $\bm{g}_{j}^{i} = \frac{\partial \lambda_{i,3}}{\partial \bm{x}_{j}}$,  $\mathbf{H}_{jj}^{i} = \frac{\partial^{2} \lambda_{i,3}}{\partial \bm{x}_{j}^{2}}$, and $\mathbf{H}_{jk}^{i} = \frac{\partial^{2} \lambda_{i,3} }{\partial \bm{x}_{j} \partial \bm{x}_{k}}$. Theorem \ref{theorem:g_H} provides their formulas. From Theorem \ref{theorem:g_H}, we know that the partial derivatives of $a_{i}$, $b_{i}$, and $c_{i}$ with respect to $\bm{x}_{j}$ and $\bm{x}_{k}$ in (\ref{equ:d_abc})  are crucial. As $a_{i}$, $b_{i}$, and $c_{i}$ are polynomials with respect to $m_{ef}$ ($e=1,2,3$ and $f=1,2,3$), to get the partial derivatives of $a_{i}$, $b_{i}$, and $c_{i}$ with respect to $\bm{x}_{j}$ and $\bm{x}_{k}$, we need to compute the partial derivatives of $m_{ef}$ with respect to $\bm{x}_{j}$ and $\bm{x}_{k}$. Section \ref{sec:d_M_i} proves these partial derivatives of $m_{ef}$. } \label{fig:alg_summary} 
\end{figure}

\begin{proof}
	
	Expanding the definition of $\bm{\Delta}_{j}^{i}$ in (\ref{equ:d_abc}), we have
	\begin{equation} \label{equ:Delta}
		\bm{\Delta}_{j}^{i} = \begin{bmatrix}
			\frac{\partial a_{i}}{\partial x_{j1}} & \frac{\partial b_{i}}{\partial x_{j1}} & \frac{\partial c_{i}}{\partial x_{j1}} \\
			\vdots & \vdots & \vdots \\
			\frac{\partial a_{i}}{\partial x_{jm}} & \frac{\partial b_{i}}{\partial x_{jm}} & \frac{\partial c_{i}}{\partial x_{jm}} \\
			\vdots & \vdots & \vdots \\
			\frac{\partial a_{i}}{\partial x_{j6}} & \frac{\partial b_{i}}{\partial x_{j6}} & \frac{\partial c_{i}}{\partial x_{j6}}
		\end{bmatrix} \in \mathbb{R}^{6 \times 6}
	\end{equation}
	Substituting the definition of $\bm{\delta}_{jm}^{i}$ in (\ref{equ:delta_jm}) into (\ref{equ:Delta}), we can write $\bm{\Delta}_{j}^{i}$ as 
	\begin{equation} \label{equ:Delta_rewrite}
		\bm{\Delta}_{j}^{i} = \begin{bmatrix}
			{\bm{\delta}_{j1}^{i}}^{T} \\
			\vdots \\
			{\bm{\delta}_{jm}^{i}}^{T} \\
			\vdots \\
			{\bm{\delta}_{j6}^{i}}^{T}
		\end{bmatrix}
	\end{equation}
	Assume $x_{jm}$ is the $m$th variable of $\bm{x}_{j}$. Then $\bm{g}_{j}^{i}$ can be written as
	\begin{equation}
		\bm{g}_{j}^{i} = \frac{\partial \lambda_{i,3}}{\partial \bm{x}_{j}} = \begin{bmatrix}
			\frac{\partial \lambda_{i,3}}{\partial {x}_{j1}} \\
			\vdots \\
			\frac{\partial \lambda_{i,3}}{\partial {x}_{jm}} \\
			\vdots \\
			\frac{\partial \lambda_{i,3}}{\partial {x}_{j6}}
		\end{bmatrix} \in \mathbb{R}^{6}.
	\end{equation}
	Here $\frac{\partial \lambda_{i,3}}{\partial x_{jm}}$ is the $m$th element of $\bm{g}_{j}^{i}$.  Substituting (\ref{equ:Delta_rewrite}) into $\bm{g}_{j}^{i}$ in (\ref{equ:g_H}), we can obtain the formula of $\frac{\partial \lambda_{i,3}}{\partial x_{jm}}$ as
	\begin{equation}
		\frac{\partial \lambda_{i,3}}{\partial x_{jm}} = -\varphi_{i}{\bm{\delta}_{jm}^{i}}^{T}\bm{\chi}_{i} = -\varphi_{i}{\bm{\delta}_{jm}^{i}}\cdot \bm{\chi}_{i}.
	\end{equation}
	
	The above formula is what we proved in Lemma \ref{lemma:first_order}. Using Lemma \ref{lemma:first_order}, we known that the formula of $\bm{g}_{j}^{i}$ in (\ref{equ:g_H}) is correct.
	
	Now we consider the Hessian matrix. According to the definitions of $\bm{\delta}_{kn}^{i}$, $\bm{\chi}_{i}$, and $\bm{\delta}_{jm}^{i}$, we have
	\begin{equation} \label{equ:delta_chi_delta_2}
		\small
		\begin{split}
			\bm{\delta}_{kn}^{i} &= \begin{bmatrix}
				\frac{\partial a_{i}}{\partial x_{kn}} \\
				\frac{\partial b_{i}}{\partial x_{kn}} \\
				\frac{\partial c_{i}}{\partial x_{kn}} 
			\end{bmatrix}, 
			\frac{\partial \bm{\chi}_{i}}{\partial x_{j,m}} = \begin{bmatrix}
				2\lambda_{i,3}^{2}\frac{\partial \lambda_{i,3}}{\partial x_{j,m}} \\
				\frac{\partial \lambda_{i,3}}{\partial x_{j,m}}\\
				0
			\end{bmatrix} ,
			\frac{\partial \bm{\delta}_{jm}^{i}}{\partial x_{kn}} = \begin{bmatrix}
				\frac{\partial^{2} a_{i}}{\partial x_{jm} \partial x_{kn}} \\
				\frac{\partial^{2} b_{i}}{\partial x_{jm} \partial x_{kn}} \\
				\frac{\partial^{2} c_{i}}{\partial x_{jm} \partial x_{kn}}
			\end{bmatrix} 
		\end{split}
	\end{equation}
	In addition, using the definition of $\varphi_{i}$ in (\ref{equ:first_order}), we obtain
	\begin{equation}\label{equ:d_inv_phi}
		\begin{split}
			\frac{\partial \varphi_{i}^{-1}}{\partial x_{km}} &= \bm{\chi}_{i} \cdot \frac{\partial \bm{\kappa}_{i}}{\partial x_{km}} + \bm{\kappa}_{i} \cdot \frac{\partial \bm{\chi}_{i}}{\partial x_{km}}\\
			& = 2\lambda_{i,3}\frac{\partial a_{i}}{\partial x_{kn}} + \frac{\partial b_{i}}{\partial x_{kn}} + \left( 2a-6\lambda_{i,3}\right) \frac{\partial \lambda_{i,3}}{\partial x_{kn}}
		\end{split}
	\end{equation}
	Let us denote the entry in the $m$th row and $n$th column  of $\mathbf{H}_{jk}^{i}$ as $\mathbf{H}_{jk}^{i}(m,n)$. Using the formula of $\mathbf{H}_{jk}^{i}$ in (\ref{equ:g_H}) and the variables in (\ref{equ:delta_chi_delta_2}) and (\ref{equ:d_inv_phi}), we have

	\begin{equation} \label{equ:H_jk(m,n)}
		\small
		\begin{split}
			\mathbf{H}_{jk}^{i}(m,n) = &-\underbrace{\varphi_{i}\frac{\partial \lambda_{i,3}}{\partial x_{kn}}\left(2\lambda_{i,3}\frac{\partial a_{i}}{\partial x_{jm}} + \frac{\partial b_{i}}{\partial x_{jm}}\right)}_{\varphi_{i}\bm{\delta}_{kn}^{i} \cdot \frac{\partial \bm{\chi}_{i}}{\partial x_{jm}}} - \\
			&\underbrace{\varphi_{i}\left(\lambda_{i,3}^{2}\frac{\partial^{2} a_{i}}{\partial x_{jm}\partial x_{kn}}+
				\lambda_{i,3}\frac{\partial^{2} b_{i}}{\partial x_{jm}\partial x_{kn}} +
				\frac{\partial^{2} c_{i}}{\partial x_{jm}\partial x_{kn}}\right)}_{\varphi_{i}\bm{\chi}_{i}\cdot\frac{\partial \bm{\delta}_{jm}^{i}}{\partial x_{kn}}} + \\
			&\underbrace{\varphi_{i}\frac{\partial \lambda_{i,3}}{\partial x_{jm}}\left( 2\lambda_{i,3}\frac{\partial a_{i}}{\partial x_{kn}} + \frac{\partial b_{i}}{\partial x_{kn}} + \left( 2a-6\lambda_{i,3}\right) \frac{\partial \lambda_{i,3}}{\partial x_{kn}}\right)}_{\varphi_{i}\frac{\partial \lambda_{i,3}}{\partial x_{jm}}\frac{\partial \varphi_{i}^{-1}}{\partial x_{km}}} \\
			= & -\varphi_{i}\left( \bm{\delta}_{jm}^{i} \cdot \frac{\partial \bm{\chi}_{i}}{\partial x_{kn}} + \bm{\chi}_{i}\cdot\frac{\partial \bm{\delta}_{jm}^{i}}{\partial x_{kn}} -\frac{\partial \lambda_{i,3}}{\partial x_{jm}}\frac{\partial \varphi_{i}^{-1}}{\partial x_{km}} \right) 
		\end{split}
	\end{equation}
	
	On the other hand, we know
	\begin{equation}
		\mathbf{H}_{jk}^{i}(m,n) = \frac{\partial^2 \lambda_{i,3}}{\partial x_{jm} \partial x_{kn}}.
	\end{equation}
	Comparing (\ref{equ:second_order}) and (\ref{equ:H_jk(m,n)}), we know that the formula of $\mathbf{H}_{jk}^{i}$ in (\ref{equ:g_H})  is correct.
\end{proof}

\subsection{Proof of Lemma \ref{lemma:mean_pt_decomp}}

\begin{proof}
	For  $j\in obs(\bm{\pi}_{i})$ and $k\in obs(\bm{\pi}_{i})$, we take $\frac{1}{N_{i}}\mathbf{T}_{j}\tilde{\bm{p}}_{ij}$ and $\frac{1}{N_{i}}\mathbf{T}_{k}\tilde{\bm{p}}_{ik}$ out  the summation. Then, we can write $\bar{\bm{p}}_{i}$ as 
	\begin{equation} \label{equ:mean_jk}
		\begin{split}
			\bar{\bm{p}}_{i}(\bm{x}_{j}, \bm{x}_{k}) &=  \mathbf{T}_{j}\underbrace{\frac{1}{N_{i}}\tilde{\bm{p}}_{ij}}_{\bm{q}_{ij}} + \mathbf{T}_{k}\underbrace{\frac{1}{N_{i}}\tilde{\bm{p}}_{ik}}_{\bm{q}_{ik}} + \underbrace{\frac{1}{N_{i}}\sum_{n \in \mathbb{O}_{jk}} \mathbf{T}_{n}\tilde{\bm{p}}_{in}}_{\bm{c}_{ijk}}. \\
			& = \mathbf{T}_{j}\bm{q}_{ij} + \mathbf{T}_{k}\bm{q}_{ik} + \bm{c}_{ijk}.
		\end{split}
	\end{equation}
	For $j\in obs(\bm{\pi}_{i})$, we can write (\ref{equ:mean_jk}) as
	\begin{equation} \label{equ:mean_j}
		\begin{split}
			\bar{\bm{p}}_{i}(\bm{x}_{j}, \bm{x}_{k}) & = \mathbf{T}_{j}\bm{q}_{ij} + \underbrace{\mathbf{T}_{k}\bm{q}_{ik} + \bm{c}_{ijk}}_{\bm{c}_{ij}} \\
			& =  \mathbf{T}_{j}\bm{q}_{ij} + \bm{c}_{ij}
		\end{split}
	\end{equation}
\end{proof}

\subsection{Proof of Theorem \ref{theorem:M_i}}

\begin{proof}
	Substituting (\ref{equ:mean_j}) into (\ref{equ:M_i}) and using the formula of $\mathbf{S}_{ij}=\mathbf{T}_{j}\mathbf{U}\mathbf{T}_{j}^{T}$ in (\ref{equ:S_p}), we have
	\begin{equation*}
		\small
		\begin{split}
			\mathbf{M}_{i}(\bm{x}_{i})  = & \sum_{j \in obs(\bm{\pi}_{i})}\mathbf{S}_{ij} - N_{i}(\mathbf{T}_{j}\bm{q}_{ij}+\bm{c}_{ij})(\mathbf{T}_{j}\bm{q}_{ij}+\bm{c}_{ij})^{T} \\
			= & \mathbf{S}_{ij}-\mathbf{T}_{j}\left( N_{i}\bm{q}_{ij}\bm{q}_{ij}^{T}\right) \mathbf{T}_{j}^{T} -\mathbf{T}_{j}\underbrace{\left(N_{j} \bm{q}_{ij}\bm{c}_{ij}^{T}\right)}_{-\mathbf{K}_{j}^{i}}  - \\
			&\underbrace{\left( N_{j}\bm{c}_{ij}\bm{p}^{T}_{ij}\right)}_{\left( -\mathbf{K}_{j}^{i}\right) ^{T}}\mathbf{T}_{j}^{T} + 
			\underbrace{\sum_{\substack{n \in obs(\bm{\pi}_{i}) \\ n \neq j}}\mathbf{S}_{in} - N_{j}\bm{c}_{ij}\bm{c}_{ij}^{T}}_{\mathbf{C}_{j}^{i}} \\
			= & \mathbf{T}_{j}\mathbf{U}_{ij}\mathbf{T}_{j}^{T} - \mathbf{T}_{j}\left( N_{i}\bm{q}_{ij}\bm{q}_{ij}^{T}\right)\mathbf{T}_{j}^{T} + \mathbf{T}_{j}\mathbf{K}_{j}^{i} + \left( \mathbf{K}_{j}^{i}\right)^{T}\mathbf{T}_{j}^{T} + \mathbf{C}_{j}^{i} \\
			= & \mathbf{T}_{j}\underbrace{\left( \mathbf{U}_{ij} - N_{j}\bm{q}_{ij}\bm{q}_{ij}^{T}\right)}_{\mathbf{Q}_{j}^{i}}\mathbf{T}_{j}^{T} + \mathbf{T}_{j}\mathbf{K}_{j}^{i} + \left( \mathbf{K}_{j}^{i}\right)^{T}\mathbf{T}_{j}^{T} + \mathbf{C}_{j}^{i} \\
			= & \mathbf{T}_{j}\mathbf{Q}_{j}^{i}\mathbf{T}_{j}^{T} + \mathbf{T}_{j}\mathbf{K}_j^{i} + (\mathbf{K}_j^{i})^{T}\mathbf{T}_{j}^{T} + \mathbf{C}_{j}^{i}
		\end{split}
	\end{equation*}
	Thus we get the formula of $\mathbf{M}_{i}(\bm{x}_{j})$ in (\ref{equ:Mij}).
	Now let us prove (\ref{equ:M_ijk}). Let us define
	\begin{equation}
		\begin{split}
			\mathbf{E}_{j}^{i} &= \mathbf{T}_{j}\bm{q}_{ij}\left( \mathbf{T}_{j}\bm{q}_{ij} + \bm{c}_{ijk}\right)^{T} \\
			\mathbf{E}_{k}^{i} &= \mathbf{T}_{k}\bm{q}_{ik}\left( \mathbf{T}_{k}\bm{q}_{ik} + \bm{c}_{ijk}\right)^{T} 
		\end{split}
	\end{equation}
	Substituting (\ref{equ:mean_jk}) into (\ref{equ:M_i}), we obtain
	\begin{equation}
		\begin{split}
			\mathbf{M}_{i}(\bm{x}_{j}, \bm{x}_{k}) =& \mathbf{T}_{j}\underbrace{\left(-N\bm{q}_{ij}\bm{q}_{ik}^{T}\right)}_{\mathbf{O}_{j}^{i}}\mathbf{T}_{k}^{T} +\mathbf{T}_{j}\underbrace{\left(-N\bm{q}_{ik}\bm{q}_{ij}^{T}\right)}_{\left( \mathbf{O}_{j}^{i}\right)^{T}}\mathbf{T}_{k}^{T} + \\
			&\underbrace{\sum_{j \in obs(\bm{\pi}_{i})}\mathbf{S}_{ij} - N_{i}\left( \mathbf{E}_{j}^{i} + \mathbf{E}_{k}^{i} + 
				\bm{c}_{ijk}\bar{\bm{p}}_{i}(\bm{x}_{j},\bm{x}_{k})^{T}\right) }_{\mathbf{C}_{jk}^{i}} \\
			= & \mathbf{T}_{j}\mathbf{O}_{jk}^{i}\mathbf{T}_{k}^{T}  +  \mathbf{T}_{k}{(\mathbf{O}_{jk}^{i})}^{T}\mathbf{T}_{j}^{T} + \mathbf{C}_{jk}^{i}
		\end{split}
	\end{equation}
	Thus we get the formula of $\mathbf{M}_{i}(\bm{x}_{j}, \bm{x}_{k})$ in (\ref{equ:M_ijk}).
\end{proof}

\subsection{Partial Derivatives of Entries in \textbf{M}$_{i}$} \label{sec:d_M_i}
As illustrated in Fig.~\ref{fig:alg_summary}, the derivatives of $a_{i}$, $b_{i}$ and $c_{i}$ in (\ref{equ:d_abc}) are the crux to get $\bm{g}_{j}^{i}$, $\mathbf{H}_{jj}^{i}$, and $\mathbf{H}_{jk}^{i}$. The $a_{i}$, $b_{i}$ and $c_{i}$ in (\ref{equ:cub}) are first-, second-, and third-order polynomials with respect to the elements in $\mathbf{M}_{i}$, respectively.  Let us denote the $e$th row and $f$th column entry of $\mathbf{M}_{i}$ as $m_{ef}$.  According to the  chain rule in calculus, to compute the partial derivatives in (\ref{equ:d_abc}), we have to calculate
\begin{equation} \label{equ:d_m}
	\frac{\partial m_{ef}}{\partial \bm{x}_{j}}, \frac{\partial^{2} m_{ef}}{\partial \bm{x}_{j}^{2}}, \ \text{and} \ \frac{\partial^{2} m_{ef}}{\partial \bm{x}_{j} \partial \bm{x}_{k}}.
\end{equation}

From our paper, we know that we only need to compute their value at $\bm{x}_{0} = [0;0;0;0;0;0]$. Assume ${q}_{ef}$, ${k}_{ef}$, and $o_{ef}$ are the $e$th row and $f$th column entries of $\mathbf{Q}_{j}^{i}$, $\mathbf{K}_{j}^{i}$, and $\mathbf{O}_{jk}^{i}$, respectively. Then, the values of $\frac{\partial m_{ef}}{\partial \bm{x}_{j}}$,  $\frac{\partial^{2} m_{ef}}{\partial \bm{x}_{j} \partial \bm{x}_{k}}$, and $\frac{\partial^{2} m_{ef}}{\partial \bm{x}_{j}^{2}}$ at $\bm{x}_{0}$ have the forms in Table \ref{table:m_ef1st-order},   Table \ref{table:2nd-cross}, and Table \ref{table:m_ef2nd-order}, respectively.

\begin{table*}
	\centering
	\begin{tabular}{  l  l }
		\toprule
		$\left. \frac{\partial m_{11}}{\partial\bm{x}_{j}}\right|_{\bm{x}_{j}=\bm{x}_{0}}  = \begin{bmatrix}
			0\\
			- 2k_{31} - 2q_{13} \\
			2k_{21} + 2q_{12} \\
			- 4k_{32} - 4q_{23} \\
			2k_{22} - 2k_{33} + 2q_{22} - 2q_{33} \\
			4k_{23} + 4q_{23} \\
		\end{bmatrix}$
		& $\left. \frac{\partial m_{12}}{\partial\bm{x}_{j}}\right|_{\bm{x}_{j}=\bm{x}_{0}} =
		\begin{bmatrix}
			4k_{31} + 4q_{13} \\
			2k_{32} + 2q_{23} \\
			2k_{33} - 2k_{11} - 2q_{11} + 2q_{33} \\
			0 \\
			- 2k_{12} - 2q_{12} \\
			- 4k_{13} - 4q_{13} \\
		\end{bmatrix}$  \\  
		$\left. \frac{\partial m_{13}}{\partial\bm{x}_{j}}\right|_{\bm{x}_{j}=\bm{x}_{0}}  =
		\begin{bmatrix}
			- 4k_{21} - 4q_{12} \\
			2k_{11} - 2k_{22} + 2q_{11} - 2q_{22} \\
			- 2k_{23} - 2q_{23} \\
			4k_{12} + 4q_{12} \\
			2k_{13} + 2q_{13} \\
			0\\
		\end{bmatrix}$ & $\left. \frac{\partial m_{22}}{\partial\bm{x}_{j}}\right|_{\bm{x}_{j}=\bm{x}_{0}}  =
		\begin{bmatrix}
			2k_{41} + 2q_{14} \\
			k_{42} + q_{24} \\
			k_{43} + q_{34} \\
			0 \\
			0 \\
			0 \\
		\end{bmatrix}$  \\  
		$\left. \frac{\partial m_{23}}{\partial\bm{x}_{j}}\right|_{\bm{x}_{j}=\bm{x}_{0}} =
		\begin{bmatrix}
			0 \\
			k_{41} + q_{14} \\
			0 \\
			2k_{42} + 2q_{24} \\
			k_{43} + q_{34} \\
			0
		\end{bmatrix}$ & $\left. \frac{\partial m_{33}}{\partial\bm{x}_{j}}\right|_{\bm{x}_{j}=\bm{x}_{0}} =
		\begin{bmatrix}
			0 \\
			0 \\
			k_{41} + q_{14} \\
			0 \\
			k_{42} + q_{24} \\
			2k_{43} + 2q_{34}
		\end{bmatrix}$   \\
		\bottomrule
	\end{tabular}
	\caption{$\frac{\partial m_{ef}}{\partial \bm{x}_{j}}$ at $\bm{x}_{j} = \bm{x}_{0}$. } \label{table:m_ef1st-order}
\end{table*}

\begin{table*}
	\small
	\centering
	\begin{tabular}{l l}
		\toprule
		$\left. \frac{\partial^{2} m_{11}}{\partial \bm{x}_{j} \partial \bm{x}_{k}}\right|_{\substack{\bm{x}_{j} = \bm{x}_{0} \\ \bm{x}_{k} = \bm{x}_{0} }} =
		\begin{bmatrix}
			0&          0&          0&          0& 0& 0 \\
			0&  8o_{33}& -8o_{32}&  4o_{34}& 0& 0 \\
			0& -8o_{23}&  8o_{22}& -4o_{24}& 0& 0 \\
			0&  4o_{43}& -4o_{42}&  2o_{44}& 0& 0 \\
			0&          0&          0&          0& 0& 0 \\
			0&          0&          0&          0& 0& 0 \\
		\end{bmatrix}$ &
		$\left. \frac{\partial^{2} m_{13}}{\partial \bm{x}_{j} \partial \bm{x}_{k}}\right|_{\substack{\bm{x}_{j} = \bm{x}_{0} \\ \bm{x}_{k} = \bm{x}_{0} }} =
		\begin{bmatrix}
			0& 4o_{23}& -4o_{22}&  2o_{24}& 0& 0 \\
			4o_{32}& - 4o_{13} - 4o_{31}&  4o_{12}& -2o_{14}& 0&  2o_{34} \\
			-4o_{22}& 4o_{21}& 0& 0& 0& -2o_{24}\\
			2o_{42}& -2o_{41}& 0& 0& 0&    o_{44}\\
			0& 0& 0& 0& 0& 0\\
			0&  2o_{43}& -2o_{42}& o_{44}& 0& 0\\
		\end{bmatrix}$  \\
		$\left. \frac{\partial^{2} m_{22}}{\partial \bm{x}_{j} \partial \bm{x}_{k}}\right|_{\substack{\bm{x}_{j} = \bm{x}_{0} \\ \bm{x}_{k} = \bm{x}_{0} }} =
		\begin{bmatrix}
			8o_{33}& 0& -8o_{31}& 0& -4o_{34}& 0 \\
			0& 0&          0& 0&          0& 0 \\
			-8o_{13}& 0&  8o_{11}& 0&  4o_{14}& 0 \\
			0& 0&          0& 0&          0& 0 \\
			-4o_{43}& 0&  4o_{41}& 0&  2o_{44}& 0 \\
			0& 0&          0& 0&          0& 0 \\
		\end{bmatrix}$&
		$\left. \frac{\partial^{2} m_{12}}{\partial \bm{x}_{j} \partial \bm{x}_{k}}\right|_{\substack{\bm{x}_{j} = \bm{x}_{0} \\ \bm{x}_{k} = \bm{x}_{0} }} =
		\begin{bmatrix}
			0& -4o_{33}& 4o_{32}& -2o_{34}& 0& 0\\
			-4o_{33}& 0& 4o_{31}&  0&  2o_{34}& 0\\
			4o_{23}&  4o_{13}& - 4o_{12} - 4o_{21}&  2o_{14}& -2o_{24}& 0\\
			-2o_{43}&  0& 2o_{41}&  0& o_{44}& 0\\
			0&  2o_{43}& -2o_{42}& o_{44}& 0& 0 \\
			0&  0& 0& 0& 0& 0 \\
		\end{bmatrix}$ 
		\\
		$\left. \frac{\partial^{2} m_{33}}{\partial \bm{x}_{j} \partial \bm{x}_{k}}\right|_{\substack{\bm{x}_{j} = \bm{x}_{0} \\ \bm{x}_{k} = \bm{x}_{0} }} =
		\begin{bmatrix}
			8o_{22}& -8o_{21}& 0& 0& 0&  4o_{24} \\
			-8o_{12}&  8o_{11}& 0& 0& 0& -4o_{14} \\
			0&          0& 0& 0& 0&          0 \\
			0&          0& 0& 0& 0&          0 \\
			0&          0& 0& 0& 0&          0 \\
			4o_{42}& -4o_{41}& 0& 0& 0&  2o_{44} \\
		\end{bmatrix}$ &
		$\left. \frac{\partial^{2} m_{23}}{\partial \bm{x}_{j} \partial \bm{x}_{k}}\right|_{\substack{\bm{x}_{j} = \bm{x}_{0} \\ \bm{x}_{k} = \bm{x}_{0} }} =
		\begin{bmatrix}
			- 4o_{23} - 4o_{32}&  4o_{31}&  4o_{21}& 0&  2o_{24}& -2o_{34} \\
			4o_{13}&          0& -4o_{11}& 0& -2o_{14}&          0 \\
			4o_{12}& -4o_{11}&          0& 0&          0&  2o_{14} \\
			0&          0&          0& 0&          0&          0 \\
			2o_{42}& -2o_{41}&          0& 0&          0&    o_{44}  \\
			-2o_{43}&          0&  2o_{41}& 0&    o_{44}&          0 \\
		\end{bmatrix} $\\
		\bottomrule
	\end{tabular}
	\caption{$\frac{\partial^{2} m_{ef}}{\partial \bm{x}_{j} \partial \bm{x}_{k}}$ at $\bm{x}_{j} = \bm{x}_{0}$ and $\bm{x}_{k} = \bm{x}_{0}$.} \label{table:2nd-cross}
\end{table*}

\begin{table*}
	\centering
	\small
	\begin{tabular}{l}
		\toprule
		$\left. \frac{\partial^{2} m_{11}}{\partial \bm{x}_{j}^{2}} \right|_{\bm{x}_{j}=\bm{x}_{0}} = 
		\begin{bmatrix}
			0 &4k_{21} + 4q_{12} & 4k_{31} + 4q_{13} & 0 & 0 & 0 \\
			4k_{21} + 4q_{12} & 8q_{33} - 8q_{11} - 8k_{11} & -8q_{23} &  4q_{34} & 0 & 0 \\
			4k_{31} + 4q_{13} & -8q_{23}& 8q_{22} - 8q_{11} - 8k_{11} & -4q_{24} & 0 & 0 \\
			0 & 4q_{34} & -4q_{24} &  2q_{44} & 0 & 0 \\
			0 & 0 & 0 & 0 & 0 & 0 \\
			0 & 0 & 0 & 0 & 0 & 0
		\end{bmatrix}$ \\
		$\left. \frac{\partial^{2} m_{12}}{\partial \bm{x}_{j}^{2}} \right|_{\bm{x}_{j}=\bm{x}_{0}}  = 
		\begin{bmatrix}
			- 4k_{21} - 4q_{12} & 2k_{11} + 2k_{22} + 2q_{11} + 2q_{22} - 4q_{33} & 2k_{32} + 6q_{23} & -2q_{34} & 0 & 0 \\
			2k_{11} + 2k_{22} + 2q_{11} + 2q_{22} - 4q_{33} & - 4k_{12} - 4q_{12} & 2k_{31} + 6q_{13} & 0 & 2q_{34} & 0 \\
			2k_{32} + 6q_{23} & 2k_{31} + 6q_{13} & - 4k_{12} - 4k_{21} - 16q_{12} &  2q_{14} & -2q_{24} & 0 \\
			-2q_{34} & 0 & 2q_{14}, 0 &   q_{44}, 0 \\
			0 & 2q_{34} &  -2q_{24} &    q_{44} & 0 & 0 \\
			0 & 0 &  0 &  0 &  0 & 0 \\
		\end{bmatrix}$ \\
		$\left. \frac{\partial^{2} m_{13}}{\partial \bm{x}_{j}^{2}} \right|_{\bm{x}_{j}=\bm{x}_{0}} = 
		\begin{bmatrix}
			- 4k_{31} - 4q_{13}& 2k_{23} + 6q_{23}& 2k_{11} + 2k_{33} + 2q_{11} - 4q_{22} + 2q_{33}&  2q_{24}& 0& 0 \\
			2k_{23} + 6q_{23}& - 4k_{13} - 4k_{31} - 16q_{13}& 2k_{21} + 6q_{12}& -2q_{14}& 0&  2q_{34} \\
			2k_{11} + 2k_{33} + 2q_{11} - 4q_{22} + 2q_{33}& 2k_{21} + 6q_{12}& - 4k_{13} - 4q_{13}& 0& 0& -2q_{24} \\
			2q_{24}& -2q_{14}& 0& 0& 0&    q_{44} \\
			0& 0& 0& 0& 0&  0 \\
			0& 2q_{34}& -2q_{24}&    q_{44}& 0& 0 \\
		\end{bmatrix}$ \\
		$\left. \frac{\partial^{2} m_{22}}{\partial \bm{x}_{j}^{2}} \right|_{\bm{x}_{j}=\bm{x}_{0}} = 
		\begin{bmatrix}
			8q_{33} - 8q_{22} - 8k_{22}& 4k_{12} + 4q_{12}& -8q_{13}& 0& -4q_{34}& 0 \\
			4k_{12} + 4q_{12}& 0& 4k_{32} + 4q_{23}& 0& 0& 0\\
			-8q_{13}& 4k_{32} + 4q_{23}& 8q_{11} - 8k_{22} - 8q_{22}& 0&  4q_{14}& 0 \\
			0& 0& 0& 0& 0& 0 \\
			-4q_{34}& 0& 4q_{14}& 0&  2q_{44}& 0 \\
			0& 0& 0& 0& 0& 0 \\
		\end{bmatrix}$ \\
		$\left. \frac{\partial^{2} m_{23}}{\partial \bm{x}_{j}^{2}} \right|_{\bm{x}_{j}=\bm{x}_{0}} = 
		\begin{bmatrix}
			- 4k_{23} - 4k_{32} - 16q_{23}& 2k_{13} + 6q_{13}& 2k_{12} + 6q_{12}& 0&  2q_{24}& -2q_{34} \\
			2k_{13} + 6q_{13}& - 4k_{32} - 4q_{23}& 2k_{22} + 2k_{33} - 4q_{11} + 2q_{22} + 2q_{33}& 0& -2q_{14}& 0 \\
			2k_{12} + 6q_{12}& 2k_{22} + 2k_{33} - 4q_{11} + 2q_{22} + 2q_{33} & - 4k_{23} - 4q_{23} & 0& 0&  2q_{14}1 \\
			0& 0& 0& 0&         0&         0 \\
			2q_{24} & -2q_{14} &  0& 0&         0&    q_{44} \\
			-2q_{34}&  0&   2q_{14}& 0&    q_{44}& 0\\
		\end{bmatrix} $\\
		$\left. \frac{\partial^{2} m_{33}}{\partial \bm{x}_{j}^{2}} \right|_{\bm{x}_{j}=\bm{x}_{0}} = 
		\begin{bmatrix}
			8q_{22} - 8k_{33} - 8q_{33}& -8q_{12}& 4k_{13} + 4q_{13}& 0& 0&  4q_{24} \\
			-8q_{12}& 8q_{11} - 8k_{33} - 8q_{33} & 4k_{23} + 4q_{23}& 0& 0& -4q_{14} \\
			4k_{13} + 4q_{13}& 4k_{23} + 4q_{23}&  0& 0& 0&  0 \\
			0& 0&  0& 0& 0&  0 \\
			0&  0&  0& 0& 0&0 \\
			4q_{24}& -4q_{14}& 0& 0& 0&  2q_{44} \\
		\end{bmatrix}$ \\
		\bottomrule
	\end{tabular}
	\caption{$\frac{\partial^{2} m_{ef}}{\partial \bm{x}_{j}^{2}}$ at $\bm{x}_{j} = \bm{x}_{0}$.} \label{table:m_ef2nd-order}
\end{table*}

{\small
\bibliographystyle{ieee_fullname}
\bibliography{second_order_PA}
}

\end{document}